\documentclass[letterpaper, 10 pt, conference]{ieeeconf}  

\IEEEoverridecommandlockouts                              

\usepackage{graphicx}
\usepackage{amssymb}
\usepackage{array}
\usepackage{multirow}
\let\labelindent\relax
\usepackage{enumitem}
\usepackage[font=small]{caption}
\usepackage[colorlinks]{hyperref}

\newcolumntype{C}[1]{>{\centering\let\newline\\\arraybackslash\hspace{0pt}}m{#1}}
\newcolumntype{L}[1]{>{\raggedright\let\newline\\\arraybackslash\hspace{0pt}}m{#1}}
\newcolumntype{R}[1]{>{\raggedleft\let\newline\\\arraybackslash\hspace{0pt}}m{#1}}

\setlist[enumerate]{wide=0pt,leftmargin=*}
\setlist[itemize]{wide=0pt,leftmargin=*}

\title{\LARGE \bf
HandoverSim: A Simulation Framework and Benchmark for Human-to-Robot Object Handovers
}

\author{
Yu-Wei Chao$^1$, Chris Paxton$^1$, Yu Xiang$^2$, Wei Yang$^1$, Balakumar Sundaralingam$^1$,
\\
Tao Chen$^{3}$, Adithyavairavan Murali$^{1}$, Maya Cakmak$^{1,4}$, Dieter Fox$^{1,4}$
\thanks{$^1$NVIDIA, $^2$UT Dallas, $^3$MIT CSAIL, $^4$University of Washington}
\thanks{$^@${\tt\small \{ychao,cpaxton,weiy,balakumars,admurali, mayacakmak,dieterf\}@nvidia.com}}
\thanks{$^@${\tt\small yu.xiang@utdallas.edu} $^@${\tt\small taochen@mit.edu}}
}

\begin{document}

\maketitle
\thispagestyle{empty}
\pagestyle{empty}

\begin{abstract}
 We introduce a new simulation benchmark ``HandoverSim'' for human-to-robot
object handovers. To simulate the giver's motion, we leverage a recent motion
capture dataset of hand grasping of objects. We create training and evaluation
environments for the receiver with standardized protocols and metrics. We
analyze the performance of a set of baselines and show a correlation with a
real-world evaluation.~\footnote{Code is open sourced at
\href{https://handover-sim.github.io}{\texttt{https://handover-sim.github.io}}.}
\end{abstract}

\section{Introduction}

The ability to exchange objects with humans seamlessly and safely is crucial
for human-robot interaction (HRI). Progress on this front can impact robots
across many application domains including domestic robots, assistive robots for
older adults and people with disabilities, and collaborative robots in
manufacturing.

Despite increasing efforts~\cite{ortenzi:tro2021}, current research on
human-robot object handovers still faces two key challenges. First, evaluation
often requires a real human in the loop. This makes the evaluation process
expensive and harder to reproduce. Second, different studies often adopt
different experimental settings (e.g., objects used) and evaluate with
different metrics. This makes cross-study comparison difficult.

Standardized datasets and benchmarks have played a key role for recent progress
in computer vision and machine
learning~\cite{lin:eccv2014,russakovsky:ijcv2015}. In robotics, there has also
been increasing efforts in improving reproducibility through standardized 
benchmarks with simulation
\cite{fan:corl2018,savva:iccv2019,yu:corl2019,xia:ral2020,james:ral2020,clegg:ral2020,erickson:icra2020,xiang:cvpr2020,deitke:cvpr2020,lin:corl2020,lee:icra2021}.
This has impacted various domains from object
manipulation~\cite{fan:corl2018,yu:corl2019},
navigation~\cite{savva:iccv2019,xia:ral2020}, to even assisting
humans~\cite{clegg:ral2020,erickson:icra2020}. Our work extends these efforts
to the critical HRI problem of object handovers.

Building a simulation-based benchmark for human-robot object handovers is
uniquely challenging. A key question is how we can simulate a realistic human
agent and its interaction with the robot during a handover process. Photo and
physics realistic simulation of humans has been widely studied in graphics but
still under active research. Furthermore, object handovers involve contact-rich
interactions between human hands and objects. A high fidelity simulation
requires substantial physics modeling and sophisticated simulation capabilities
on soft body dynamics.

In this work, we introduce HandoverSim: a new simulation framework and
benchmark for human-to-robot object handovers. We focus on the less explored
but challenging human-to-robot (H2R) paradigm~\cite{ortenzi:tro2021}, where the
robot has to take over an object handed over by a human. As the first step, we
focus specifically on the realism of hand motion in simulating the human giver.
We leverage a recent motion capture dataset of human grasping objects and
performing handover attempts~\cite{chao:cvpr2021}, and build a simulation
environment where the motion of the human giver is driven by the captured
motion. Overall, the environment contains 1,000 handover scenes captured from
10 subjects handing over 20 different objects in the real world. Based on this
environment, we create a new benchmark for training and evaluating robots on
H2R object handovers.

Our contributions are threefold. First, we propose a new framework for
simulating the giver's motion in H2R handovers. Second, we introduce a new
benchmark environment which enables standardized and reproducible evaluation of
receiver policies. To the best of our knowledge, this is the first simulation
benchmark for H2R handovers. Finally, we analyze the performance of a set of
baselines, including a motion planning method, a task planning method, and a
pre-trained reinforcement learning policy. We further show a positive
correlation between the performance achieved on our benchmark and a real-world
user study.

\section{Related Work}

\noindent \textbf{Object Handovers.} Human-robot handovers has been
increasingly studied over the past decade~\cite{ortenzi:tro2021}, achieving
impressive progress on different robot capabilities including intent
communication~\cite{newbury:arxiv2021}, grasp planning~\cite{yang:icra2021},
perception~\cite{yang:iros2020,rosenberger:ral2021}, handover location
selection~\cite{nemlekar:icra2019,ardon:ral2021}, motion planning and
control~\cite{basili:hcrs2009}, grip force modulation~\cite{mason:ebr2005} and
error handling~\cite{eguiluz:icra2017}. However, previous works diverge on
experimental settings and metrics, making fair comparisons difficult. Our
HandoverSim evaluates H2R handovers in a physics simulated environment with a
broad set of objects and a unified set of evaluation metrics commonly used in
handover research. Its introduction can facilitate easy and fair comparison
among different handover approaches.


\vspace{1mm}
\noindent \textbf{Simulation for Robotics.} Simulation environments have been
increasingly used in robotics since they enable scalable training of robots and
standardized evaluation. Some environments focus specifically on simulating
large-scale indoor
scenes~\cite{savva:arxiv2017,kolve:arxiv2017,wu:arxiv2018,yan:arxiv2018,das:cvpr2018,xia:cvpr2018,savva:iccv2019,deitke:cvpr2020}.
They are typically used for navigation related tasks and are often lacking on
interactability and physics realism. Some others focus on object manipulation
and thus require a high fidelity physics simulation for realistic
interactions~\cite{fan:corl2018,yu:corl2019,james:ral2020,xia:ral2020,xiang:cvpr2020,lin:corl2020,ehsani:cvpr2021}.
Nonetheless, neither of these environments contain simulated humans. Most
related to ours are the recently introduced Assistive
Gym~\cite{erickson:icra2020} and Watch-and-Help~\cite{puig:iclr2021}, both
contain human-like agents in simulation. Assistive Gym~\cite{erickson:icra2020}
is a physics simulated environment for training robots to assist people with
activities of daily living. While they simulate robot-human physical
interactions, their virtual humans are either completely passive or driven by
motion trained with a cost function. In contrast, we drive our simulated human
hands with motion captured from real humans.
Watch-and-Help~\cite{puig:iclr2021} is an environment based on
VirtualHome~\cite{puig:cvpr2018} for evaluating social intelligence. They focus
primarily on high-level task learning and thus do not simulate realistic
physical interactions.

\begin{figure}[t]
 \centering
 \includegraphics[width=0.959\linewidth]{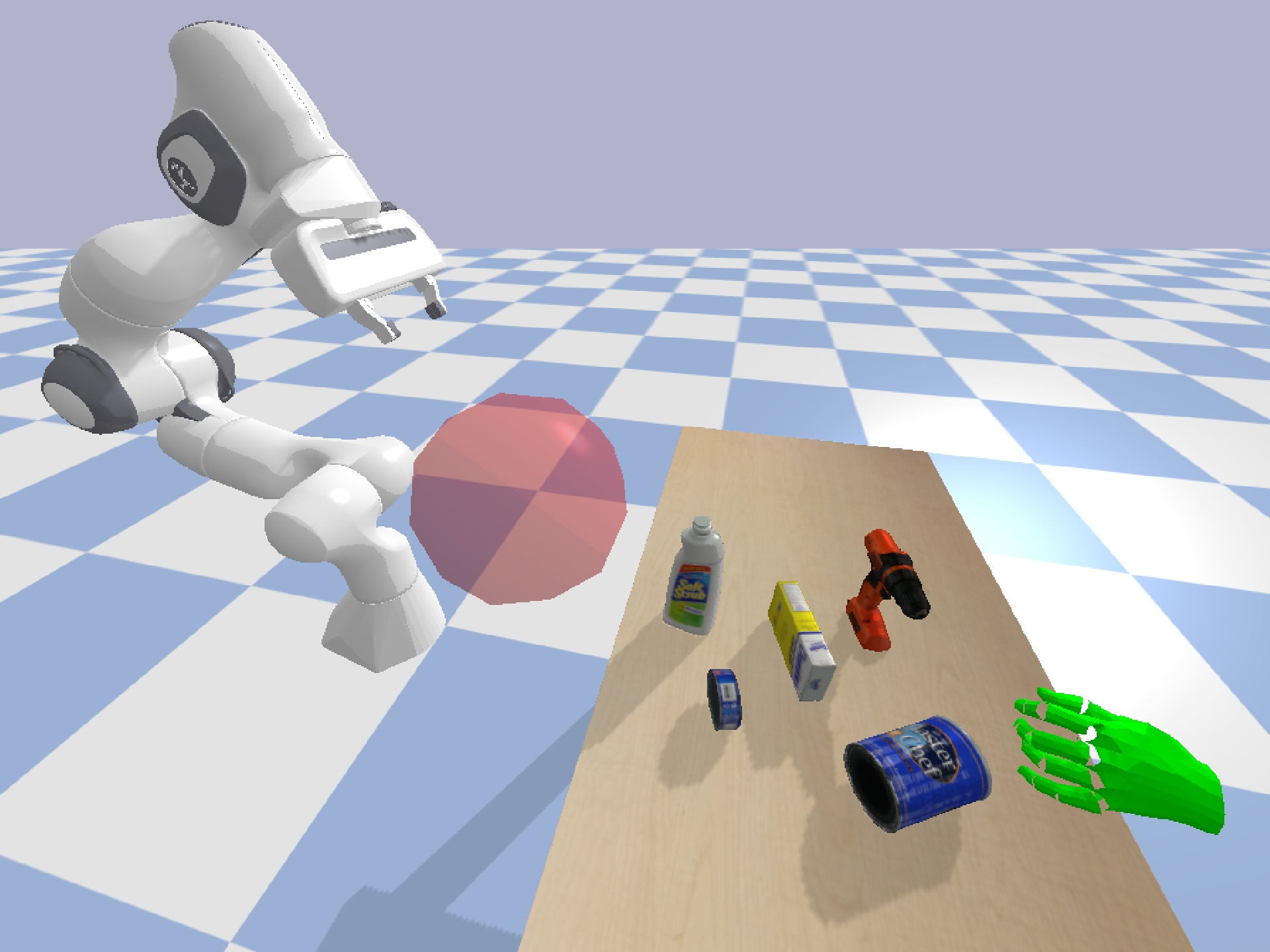}
 \caption{\small Our simulation environment for human-to-robot object
handovers. Green: human hand. Red sphere: goal region.}
 \label{fig:environment}
\end{figure}

\vspace{1mm}
\noindent \textbf{Handover Benchmarks.} Our work is also related to recent
efforts on standardizing the experimental setting and protocol for handovers.
Ye et al.~\cite{ye:iccv2021} proposed a large-scale human-to-human handover
dataset with object and hand pose annotations, and used it to study human grasp
prediction. Rather than predicting human grasps, Chao et
al.~\cite{chao:cvpr2021} studied robot grasp generation for safe H2R handovers.
While these works promote a fair benchmark for handover research, their tasks
are formulated only at the vision level, without any physics simulated
evaluation. Sanchez-Matilla et al.~\cite{sanchez-matilla:ral2020} proposed a
real-world benchmark for H2R handovers of unseen cups with unknown filling.
However, they only considered objects of a single category, i.e., cups. Our
HandoverSim contains the commonly used YCB objects~\cite{xiang:rss2018} and
allows a physics simulated evaluation of the full handover process.

\section{Simulating Handovers}

We assume the scene contains a human giver and a robot arm receiver facing each
other with a table in between, and a set of objects initially placed on the
table, as shown in Fig.~\ref{fig:environment}. The human giver will pick up an
object from the table with a single hand (right or left) and offer it to the
robot. The robot receiver is able to observe the human's actions and the scene,
and react simultaneously to eventually take over the object from the human's
hand.

To simulate the physical interaction of this process, we build a simulation
environment using the PyBullet physics
engine~\cite{coumans:2021}.~\footnote{Support of Isaac
Gym~\cite{makoviychuk:arxiv2021} has been added upon publication of the paper.}
For the choice of the robot, we use a model of the commercial Franka Emika
Panda with a 7-DoF arm and a 2-DoF parallel-jaw gripper. While the described
handover process is a two-agent game (between the human and robot), in the
benchmark we expect the robot to be the only controllable agent and can move
freely within its own physical limits. A key question is then how we can
simulate a realistic human giver, particularly on their motion and interaction
with the robot. Below we describe our approach.

\vspace{1mm}
\noindent \textbf{Grasping and Offering Object.} To simulate realistic human
motion of object grasping and offering, we leverage a recent human grasping
dataset called DexYCB~\cite{chao:cvpr2021}. DexYCB captures real motion of
human subjects picking up an object from a table of objects, and handing it to
an imagined partner across the table. A typical capture starts from the subject
in a relaxed pose, and ends in the subject's hand holding the object in the
air, waiting for a receiver to acquire it. Each capture provides frame-wise 3D
pose of both the hand in use and the objects in the scene. This data serves as
an ideal basis for the human giver's motion in the ``pre-handover''
phase~\cite{ortenzi:tro2021} (i.e., before the receiver's hand contacting the
object). We therefore use these captures to drive the human giver's motion in
simulation.

We first import object and hand models into simulation. DexYCB uses 20 objects
from the YCB-Video dataset~\cite{xiang:rss2018}, where object pose is
represented by the 6D pose of a rigid mesh. Hand pose is represented by the
deformable MANO mesh model~\cite{romero:siggraphasia2017}, parameterized by two
low-dimensional embeddings for shape and articulation. We 
use~\cite{kalevatykh:2020} to import MANO into PyBullet, which turns the hand
into an articulated rigid body after the shape deformation.

\begin{figure*}[t]
 \centering
 \includegraphics[width=0.161\linewidth]{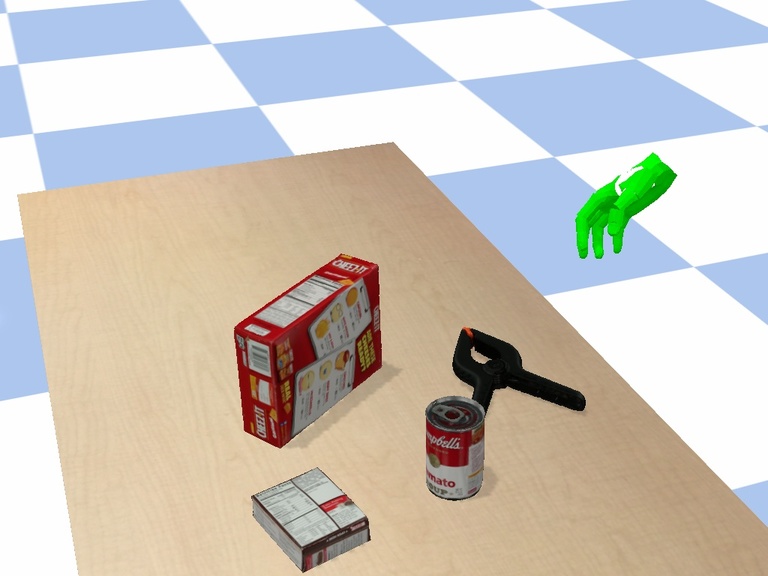}
 \includegraphics[width=0.161\linewidth]{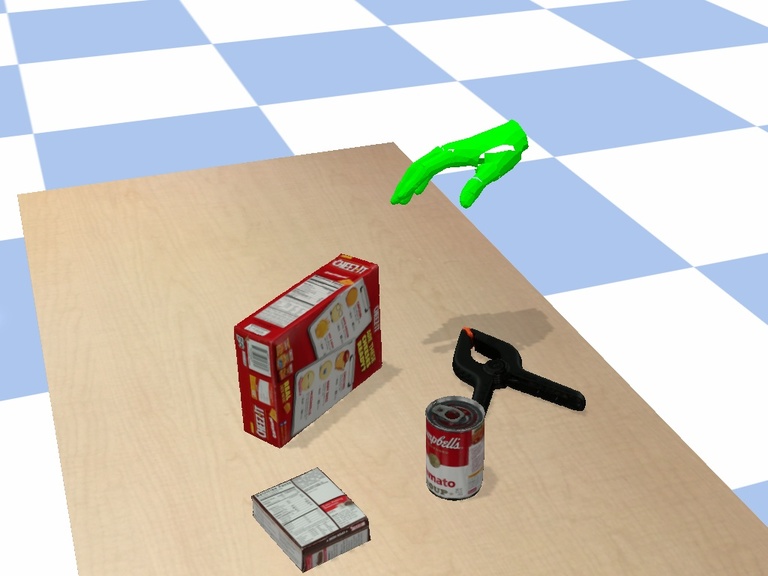}
 \includegraphics[width=0.161\linewidth]{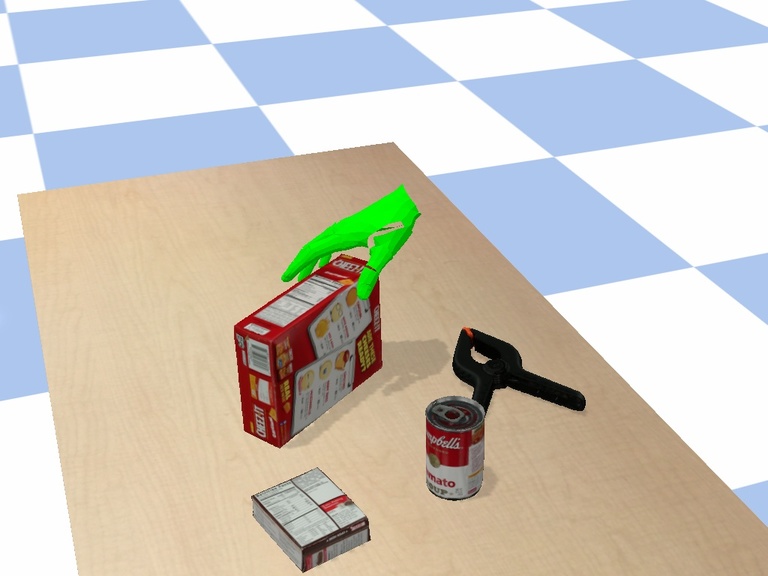}
 \includegraphics[width=0.161\linewidth]{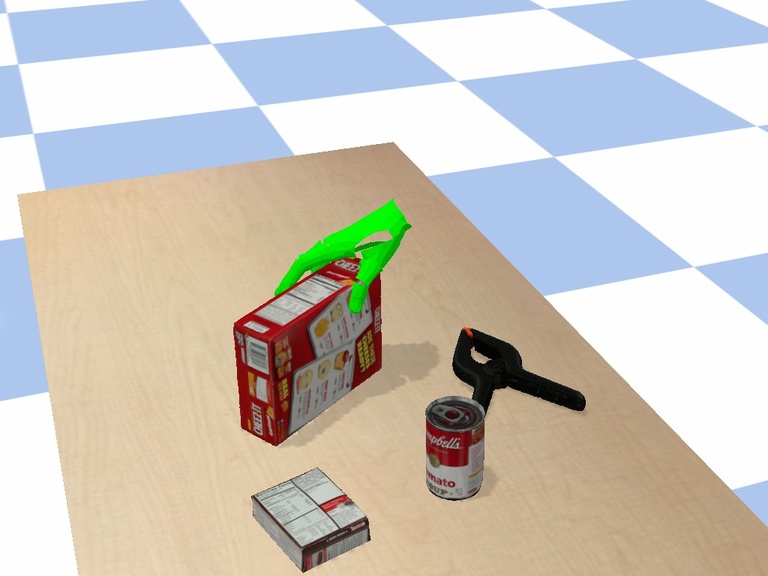}
 \includegraphics[width=0.161\linewidth]{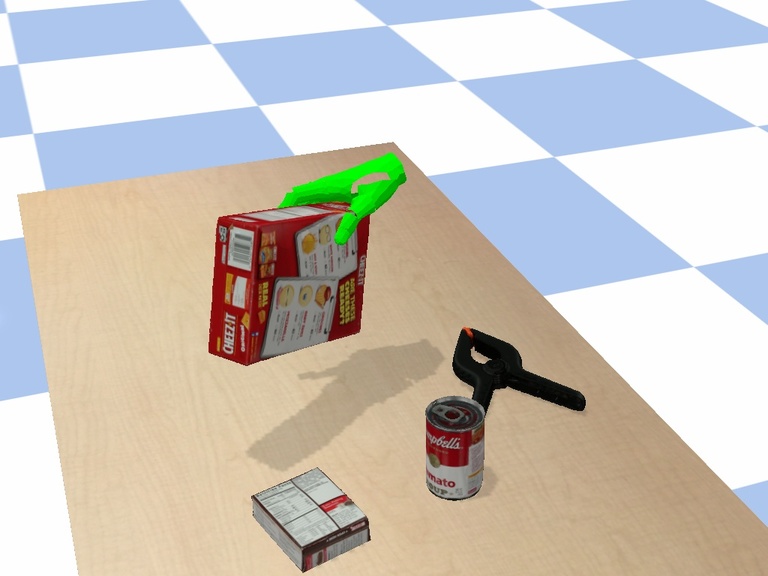}
 \includegraphics[width=0.161\linewidth]{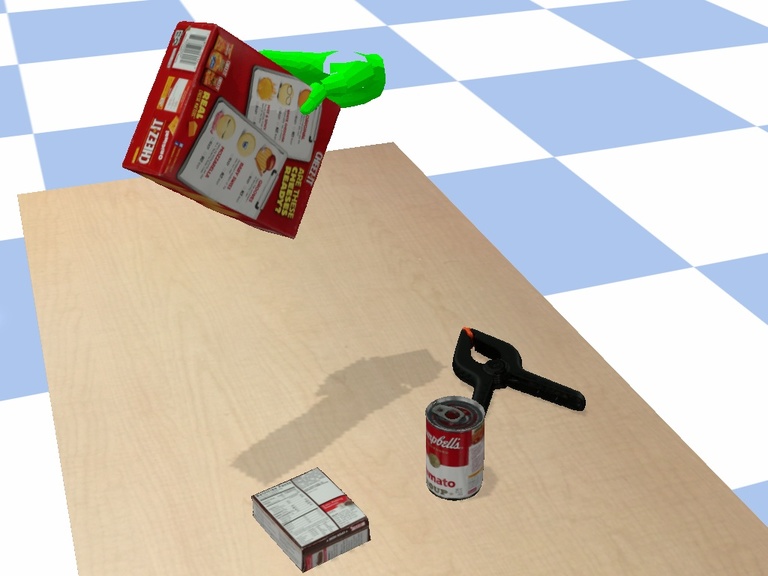}
 \\ \vspace{1mm}
 \includegraphics[width=0.161\linewidth]{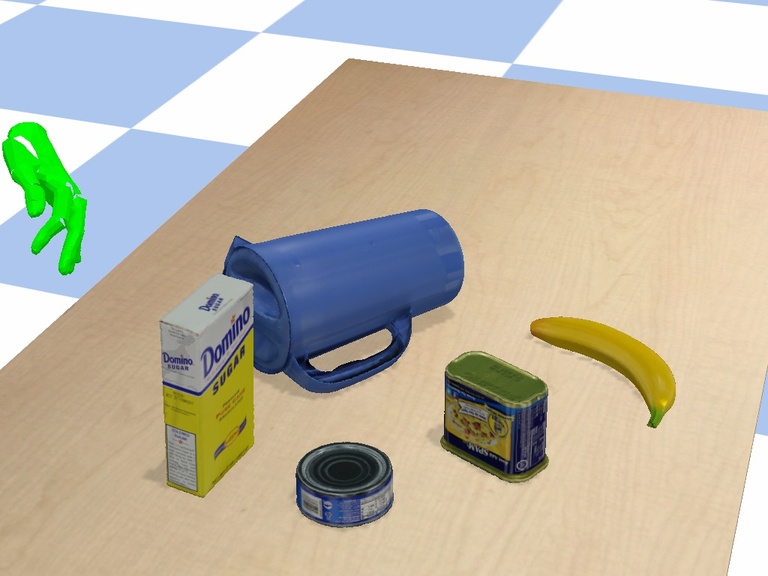}
 \includegraphics[width=0.161\linewidth]{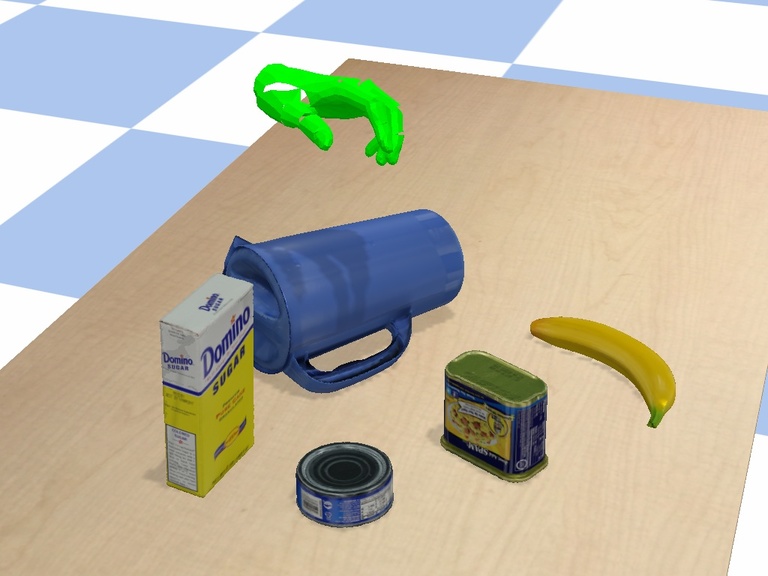}
 \includegraphics[width=0.161\linewidth]{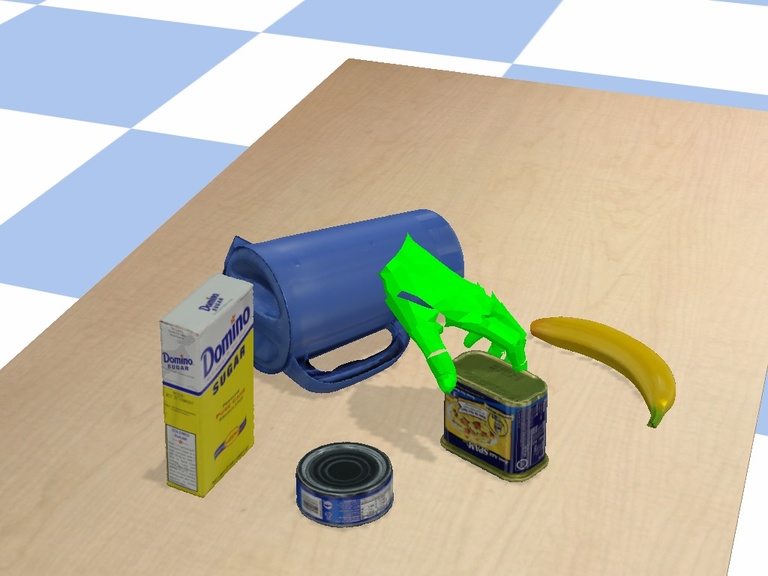}
 \includegraphics[width=0.161\linewidth]{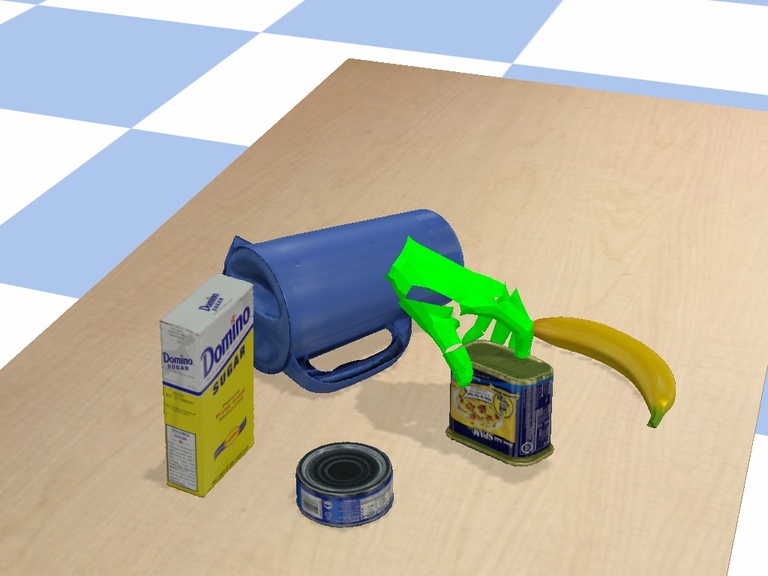}
 \includegraphics[width=0.161\linewidth]{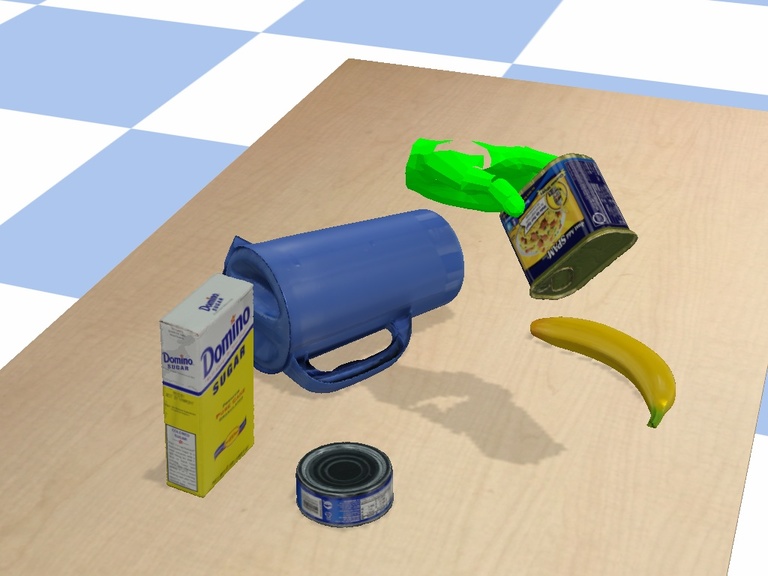}
 \includegraphics[width=0.161\linewidth]{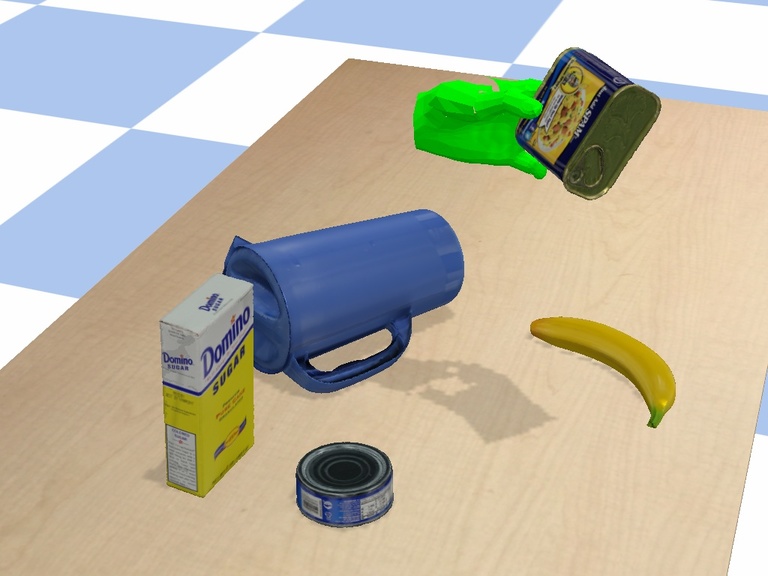}
 \caption{\small Simulated human hand and object motion for handovers. Top:
right hand. Bottom: left hand.}
 \label{fig:human}
\end{figure*}

\begin{figure}[t]
 \centering
 \includegraphics[width=0.316\linewidth, trim=170 154 404 100, clip]{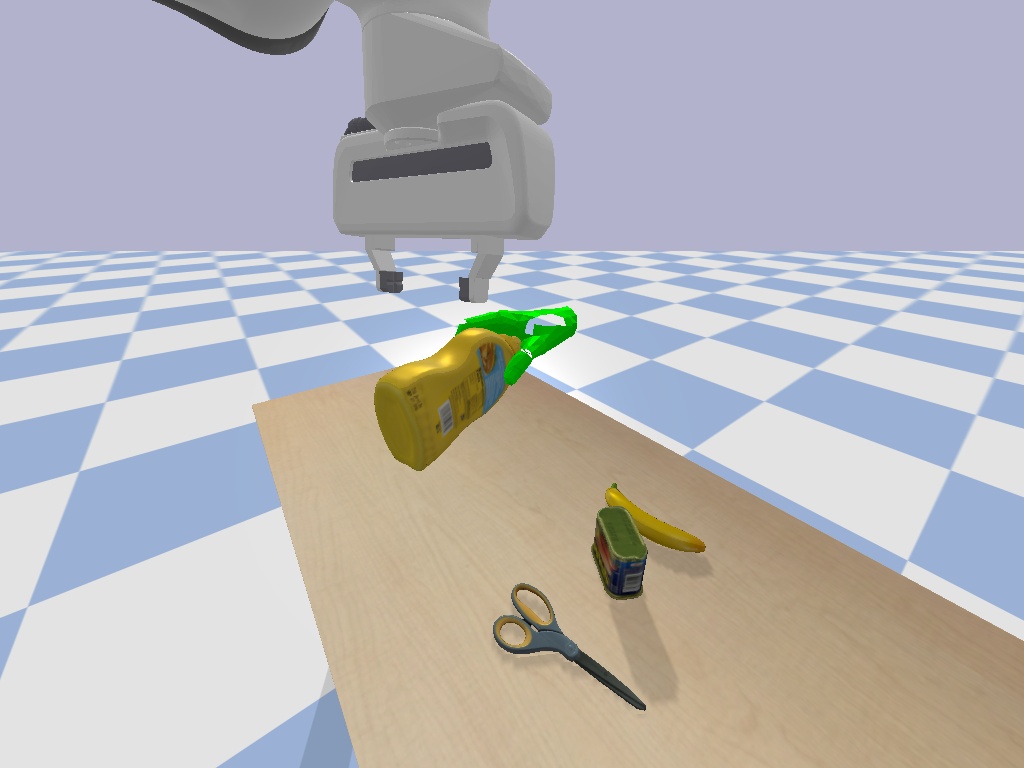}
 \includegraphics[width=0.316\linewidth, trim=170 154 404 100, clip]{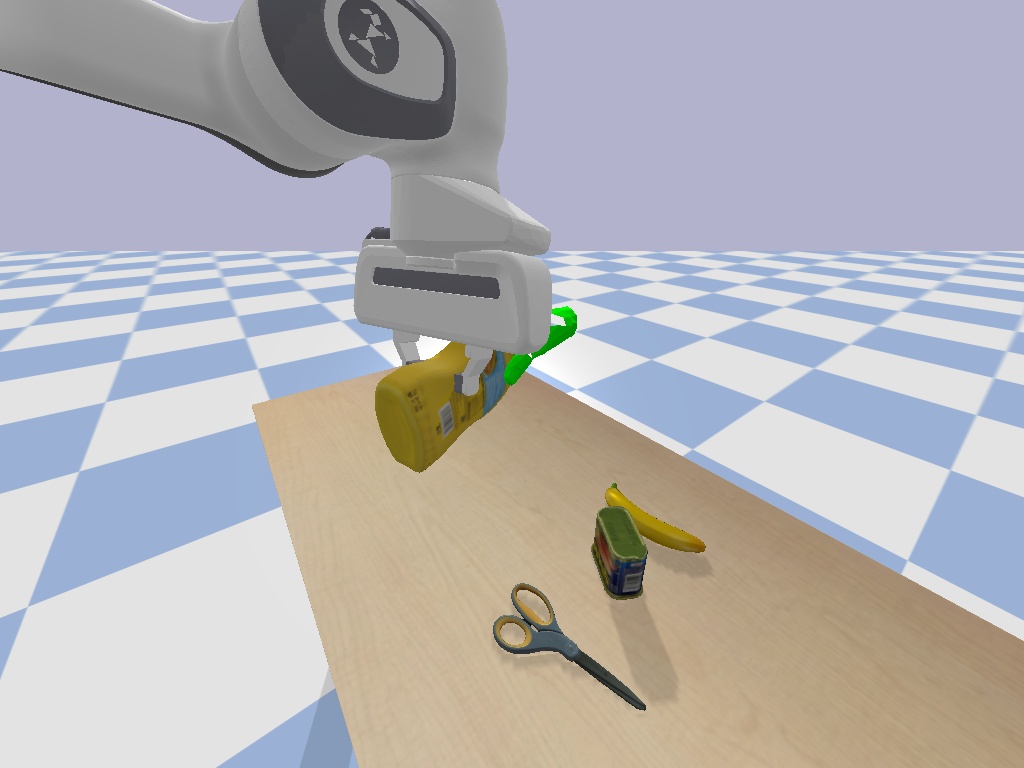}
 \includegraphics[width=0.316\linewidth, trim=170 154 404 100, clip]{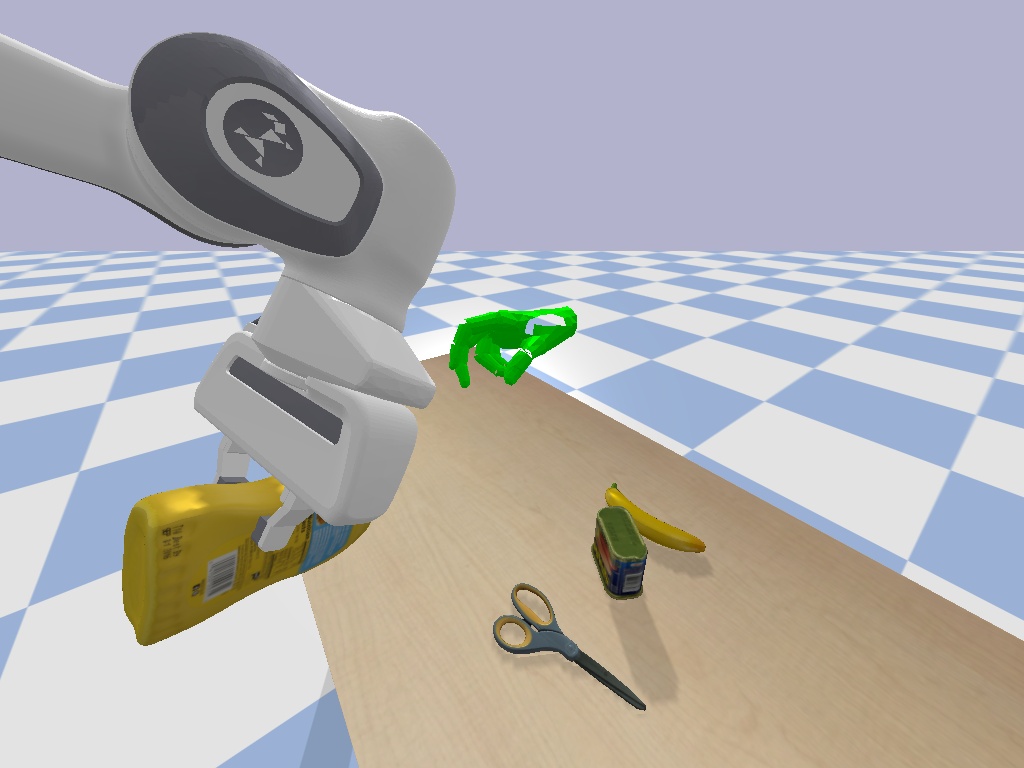}
 \\ \vspace{1mm}
 \includegraphics[width=0.316\linewidth, trim=287 154 287 100, clip]{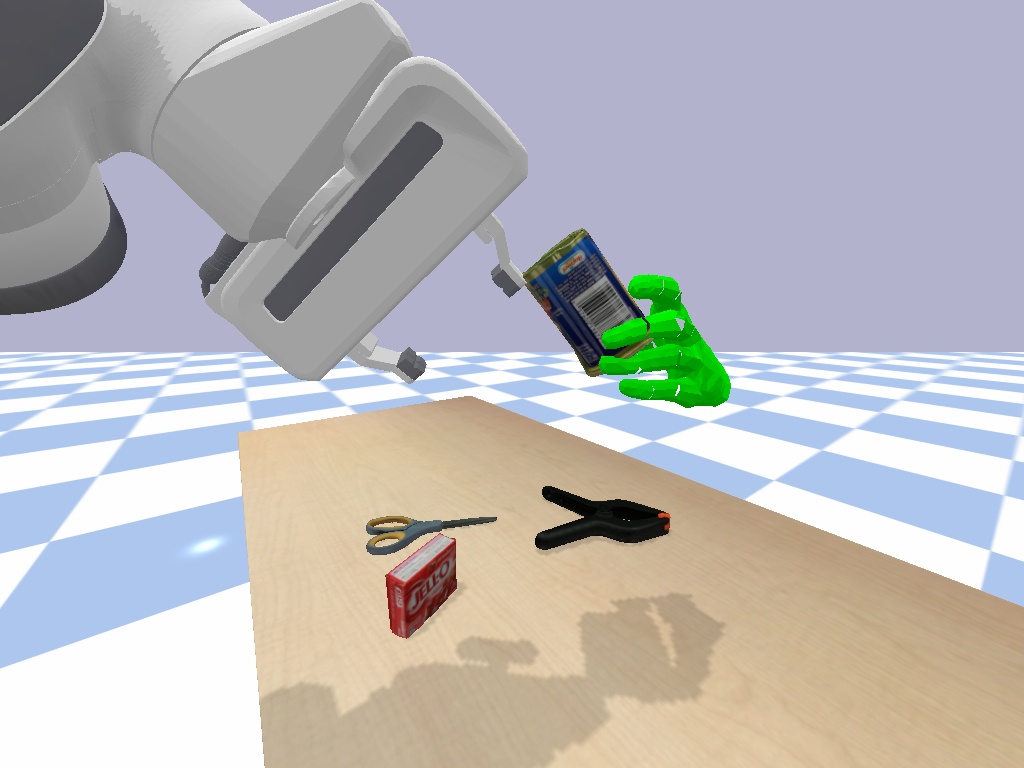}
 \includegraphics[width=0.316\linewidth, trim=287 154 287 100, clip]{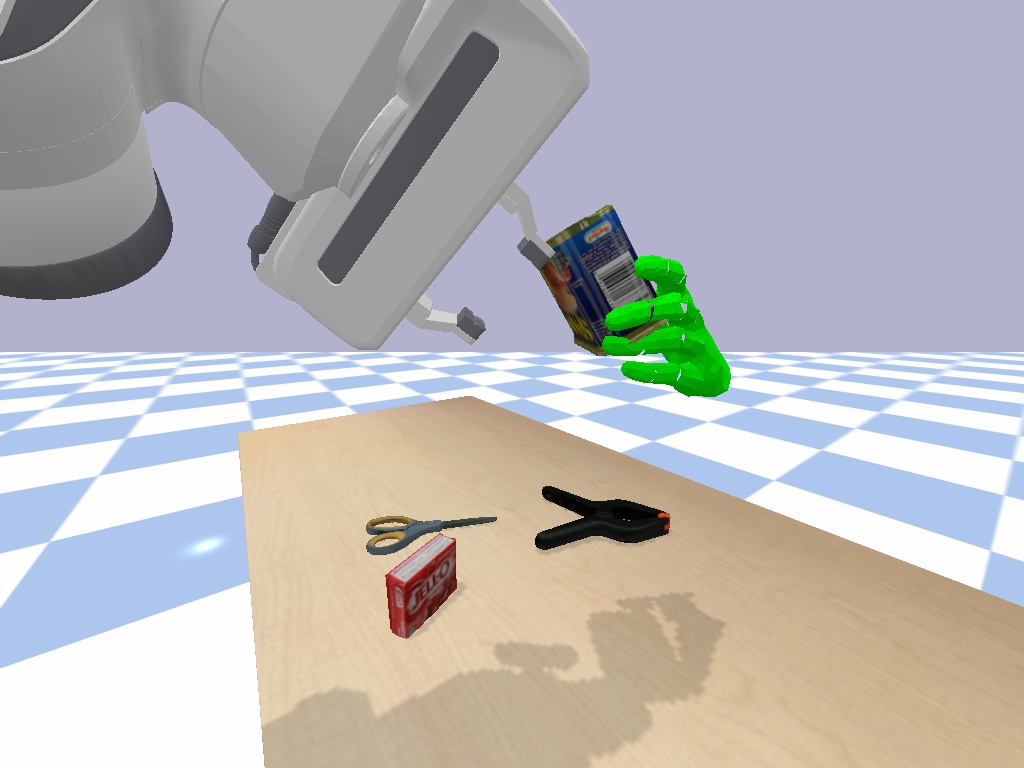}
 \includegraphics[width=0.316\linewidth, trim=287 154 287 100, clip]{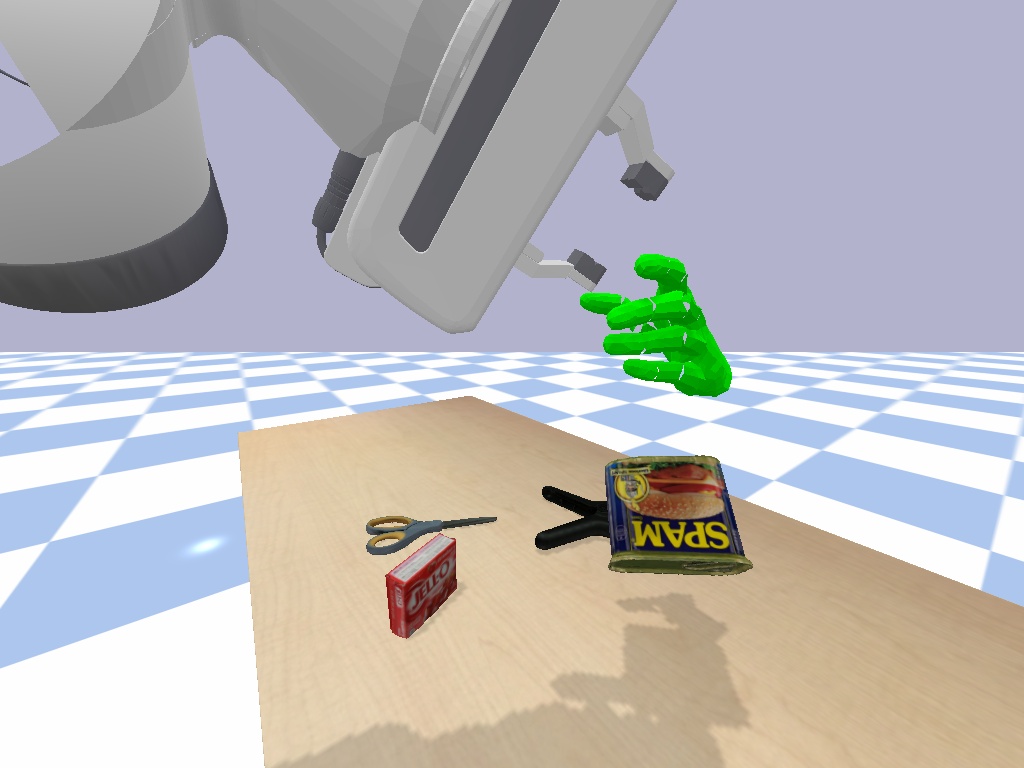}
 \caption{\small Simulating objects released from human hands. A release is
triggered at the middle frame in the top and bottom examples.}
 \label{fig:release}
\end{figure}

One approach to reproducing the captured motion in simulation is to learn a
control policy to actuate the human hand to pick up the object with close
fidelity to their captured trajectories~\cite{peng:rss2020}. However, learning
to control a dexterous hand to manipulate objects has been notoriously
challenging~\cite{rajeswaran:rss2018,openai:arxiv2018,nagabandi:corl2019} and
requires a significant engineering effort on the hand's physics model. Hence we
take a different approach: rather than relying on the human hand to physically
move the object, we augment the object models by adding additional actuators to
their base such that their 6D pose can be directly actuated by controllers in
simulation. Consequently, we can directly control the human hand and the
grasped object to simultaneously move according to their captured trajectories.
Due to the noise in motion capture, the data may contain moderate
interpenetration between the hand and object model during grasping. We thus
disable the collision detection between the human hand and object to avoid
artifacts caused by unstable simulation. For each capture, we simulate a
handover trial by ``replaying'' the motion from the first frame. After reaching
the last frame, we force the hand and object to stay in the end pose, i.e.,
simulating the giver waiting still for the receiver. Fig.~\ref{fig:human} shows
two examples of the simulated human hand and object motion.

\vspace{1mm}
\noindent \textbf{Releasing Object.} Once we can simulate a human presenting an
object for handover, the next question to simulate the release of the object to 
the robot, i.e., the ``physical handover'' phase~\cite{ortenzi:tro2021}. Since
the object's pose has so far been actively controlled, we need to disable the
controller to simulate a release. We simulate releases in two scenarios. An 
``active'' release (i.e., a voluntary release from human) is triggered when the
object has been in contact with both gripper finger's gripping surface
continuously for 0.1 seconds. In addition to releasing voluntarily, the object
can also drop involuntarily, e.g., due to robot arm strikes. Therefore, we also
trigger a ``passive'' (i.e., compulsive) release when the object has been free
of contact with either gripping surface but has been in contact with any other
parts of the robot body continuously for 0.1 seconds. Once triggered, the
object becomes completely passive and is only subject to robot forces and
gravity. Fig.~\ref{fig:release} shows two examples of the triggered release.

\vspace{1mm}
\noindent \textbf{Remarks.} We make two remarks regarding our simulation of the
human giver: (1) Our human's motion is non-adaptive to the robot's actions.
This assumption is rather naive but still applies in many human-robot
collaboration scenarios where the human is cognitively engaged in some other
tasks and has to exchange objects with robots ``blindly''. (2) We do not
simulate realistic human hand motion during object release and post-release due
to the lack of such data.

%
%
%

\section{Benchmark Environment}
\label{sec:benchmark}

Using the proposed simulation framework, we construct a new benchmark
environment called HandoverSim.

\vspace{1mm}
\noindent \textbf{Task.} We formalize the task as standard reinforcement
learning (RL) problems and implement the benchmark environment using the OpenAI
Gym API~\cite{brockman:arxiv2016}. At each time step $t$, the robot agent
observes the current state of the environment $s_t\in\mathcal{S}$, and has to
generate an action $a_t\in\mathcal{A}$ from its policy. The action $a_t$ is
then executed in the environment, which returns a new state $s_{t+1}$ and
optionally a scalar reward $r$ if an RL algorithm is used. $\mathcal{A}$ is a
9-dimensional continuous space where an action specifies the target joint
position for a PD controller of the robot arm (7 DoF) and gripper (2 DoF).
$\mathcal{S}$ may vary depending on the benchmark setting, e.g., object pose or
point cloud.

\begin{table*}[t]
 \centering
 \small
 \begin{tabular}{l|cccc|cccc|cccc|cccc}
  \hline
        & \multicolumn{4}{c|}{S0: default} & \multicolumn{4}{c|}{S1: unseen subjects} & \multicolumn{4}{c|}{S2: unseen handedness} & \multicolumn{4}{c}{S3: unseen grasping} \\
        & \#sub & hand & \#obj & \#sce & \#sub & hand & \#obj & \#sce & \#sub & hand & \#obj & \#sce & \#sub & hand & \#obj & \#sce \\
  \hline
  train & 10 & R/L & 18 & ~~720 & ~7 & R/L & 18 & ~~630 & 10 & R   & 18 & ~~449 & 10 & R/L & 14 & ~~700 \\
  val   & ~2 & R/L & 18 & ~~~36 & ~1 & R/L & 18 & ~~~90 & ~2 & L   & 18 & ~~~91 & 10 & R/L & ~2 & ~~100 \\
  test  & ~8 & R/L & 18 & ~~144 & ~2 & R/L & 18 & ~~180 & ~8 & L   & 18 & ~~360 & 10 & R/L & ~2 & ~~100 \\
  \hline
  all   & 10 & R/L & 18 & ~~900 & 10 & R/L & 18 & ~~900 & 10 & R/L & 18 & ~~900 & 10 & R/L & 18 & ~~900 \\
  all*  & 10 & R/L & 20 & 1,000 & 10 & R/L & 20 & 1,000 & 10 & R/L & 20 & 1,000 & 10 & R/L & 20 & 1,000 \\
  \hline
 \end{tabular}
 \caption{\small Statistics of the four evaluation setups: S0, S1, S2, and S3.
For each setup, we list the number of subjects (\#sub), hands used (right: R,
left: L, or both: R/L), number of objects (\#obj), and number of scenes
(\#sce). ``all*'' includes all the sequences from DexYCB, and ``all'' is after
removing objects ungraspable by the gripper.}
 \label{tab:setup}
\end{table*}

For quantitative evaluation we need to programmatically define when a task is
succeeded or failed. We claim that a \textit{success} is achieved when the 
below three conditions are met:
\begin{enumerate}
 \item The gripper fingers are in contact with the handed over object.
 \item The position of the gripper link lies with a pre-specified goal region
(Fig.~\ref{fig:environment}).
 \item The above two conditions hold true continuously
for 0.1 seconds.
\end{enumerate}
We establish a spherical goal region (Fig.~\ref{fig:environment}) in close proximity
in front of the robot arm to prevent a success case where the robot has reached
the object but got stuck due to an unnatural pose configuration. The robot
instead should be able to pull back the object to close proximity after taking
hold of it and potentially use it to perform other tasks. On the other hand, a
\textit{failure} is detected when either one of the following three conditions
is met:
\begin{enumerate}
 \item Any part of the robot body is in contact with any part of the human
hand.
 \item At least one of the gripper fingers is not in contact with the handed
over object and the object is in contact with the table or other objects or its
center falls below the height of the table surface.
 \item A maximum time limit of 13 seconds has reached.
\end{enumerate}
The first condition (referred to as ``contact'') prohibits robot-human contacts
to avoid potential harms to human (e.g., human hand pinched by the gripper) and
ensures a safe handover process. The second condition (referred to as ``drop'')
prohibits the robot from dropping the object. The third condition (referred to
as ``timeout'') ensures that the task is completed within a reasonable time
length.

\vspace{1mm}
\noindent \textbf{Evaluation Metrics.} Object handover is commonly regarded as
a multi-objective task in HRI~\cite{ortenzi:tro2021}. For our benchmark, we
report metrics on \textit{efficacy}, \textit{efficiency} and \textit{safety}.
First, using the definitions of success and failure mentioned above, we
terminate an episode whenever a success or failure is detected. This way an
episode can only belong to a success or failure case, but not both, i.e., we do
not allow the robot to complete the task after touching the human hand. To
evaluate efficacy, we calculate the success rate over all the test episodes,
and also the failure rate from each of the three failure causes: ``contact'',
``drop'', and ``timeout''. For efficiency, we calculate the mean completion
time over the successful episodes. Note that we do not include episodes of
failure in this metric. We further divide the completion time into execution
time and planning time. The execution time (``exec'') is the accumulated time
during which the robot is physically moving, and depends only on the number of
time steps in an episode. The planning time (``plan'') is the accumulated wall
time on running the policy function. Finally, we regard the failure rate due to
robot-human contacts (i.e., ``contact'') as our safety metric.

\vspace{1mm}
\noindent \textbf{Environment Statistics.} DexYCB~\cite{chao:cvpr2021} captures
10 subjects grasping 20 objects with 5 trials per subject-object pair. This
amounts to 1,000 motion capture sequences where each sequence captures a single
trial. We adopt all the sequences and simulate one handover scene from each
sequence, resulting in 1,000 \textit{scenes}. Among the 5 trials, the first two
use the right hand, the next two use the left hand, and the choice of the last
one is randomized. This results in approximately an equal number of handover
attempts from both the right and left hand. The object to be grasped is
initially placed on the table and mixed with 2 to 4 other objects in randomized
pose configuration. Each sequence (for simulating the ``pre-handover'' phase)
is slightly less than 3 seconds, containing the full course of action from
pickup to offering for handover.

\vspace{1mm}
\noindent \textbf{Training and Evaluation Setup.} We expect the benchmark to be
used not only for evaluation but also for training. Therefore, we divide the
scenes into train/val/test splits following standard machine learning
paradigms. Due to limitations in gripper capacity, we first remove scenes where
the robot has to grasp the following two objects: ``002\_master\_chef\_can''
and ``036\_wood\_block''. Next, following DexYCB~\cite{chao:cvpr2021}, we
generate four different \textit{setups} by splitting the scenes in four
different ways to benchmark different scenarios:
\begin{itemize}
 \item \textbf{S0 (default).} The train split contains all 10 subjects and all 18
grasped objects.
 \item \textbf{S1 (unseen subjects).} The scenes are split by subjects
(train/val/test: 7/1/2).
 \item \textbf{S2 (unseen handedness).} The scenes are split by handedness, 
i.e., right or left hand used (train/val/test: R/L/L). 
 \item \textbf{S3 (unseen grasping).} The scenes are split by the grasped
objects (train/val/test: 14/2/2).
\end{itemize}
Tab.~\ref{tab:setup} shows the statistics of the four setups. For
evaluation, each policy is ran for one single episode in each test scene.

\section{Experiments}

We select a set of baselines and studied their performance on HandoverSim. As
the first benchmark, we study a simple setting where we assume the ground-truth
states of the human hand and objects (retrieved from simulation) are available
to the policy, i.e., we assume a perfect perception and focus the evaluation
solely on planning and control capabilities. For a fair comparison of planning
time, all the baselines are ran on the same platform with an AMD Ryzen 9 5950X
CPU with 128 GB RAM and an NVIDIA GeForce RTX 3090 GPU.

\vspace{1mm}
\noindent \textbf{Baselines.} Our baselines are adopted from three prior works:
\begin{itemize}
 \item \textbf{OMG Planner~\cite{wang:rss2020}.} This is a joint motion and
grasp planner. It takes in the robot's current configuration and a set of
candidate goal configurations, and jointly selects a goal and generates a
trajectory towards it based on trajectory optimization. Since this is an open
loop planner, we force the robot to stay in the start pose while the human hand
is moving. Once the human hand reaches the end pose and begins the wait, we
then use the object's pose for planning. To obtain the goal set, the planner
uses pre-generated grasp poses for each YCB object from~\cite{eppner:isrr2019}.
Once the robot traverses to the end of the planned trajectory, we switch to a
hand coded policy that closes the gripper and moves towards the handover goal
region. Note that this baseline does not take the human hand into account.

\begin{table*}[t]
 \centering
 \small
 \setlength{\tabcolsep}{2.00pt}
 \begin{tabular}{l||cccc|ccc||cccc|ccc}
  \hline
  & \multicolumn{7}{c||}{S0: default} & \multicolumn{7}{c}{S1: unseen subjects} \\
  \hline
  & success & \multicolumn{3}{c|}{mean accum time (s)} & \multicolumn{3}{c||}{failure (\%)} & success & \multicolumn{3}{c|}{mean accum time (s)} & \multicolumn{3}{c}{failure (\%)} \\
  & (\%) & exec & plan & total & contact & drop & timeout & (\%) & exec & plan & total & contact & drop & timeout \\
  \hline
  OMG Planner~\cite{wang:rss2020}       & 62.50          & ~8.309          & ~1.414          & ~9.722          & 27.78          & ~\textbf{8.33} & ~\textbf{1.39} & \textbf{62.78} & ~8.012          & ~1.355          & ~9.366          & 33.33          & ~\textbf{2.22} & ~\textbf{1.67} \\
  Yang et al.~\cite{yang:icra2021}      & \textbf{64.58} & ~4.864          & ~\textbf{0.036} & ~4.900          & 17.36          & 11.81          & ~6.25          & \textbf{62.78} & ~4.719          & ~\textbf{0.039} & ~4.758          & 14.44          & ~7.78          & 15.00          \\
  GA-DDPG~\cite{wang:corl2021} hold     & 50.00          & ~7.139          & ~0.142          & ~7.281          & ~\textbf{4.86} & 19.44          & 25.69          & 55.00          & ~6.791          & ~0.136          & ~6.927          & ~\textbf{8.89} & 15.00          & 21.11          \\
  GA-DDPG~\cite{wang:corl2021} w/o hold & 36.81          & ~\textbf{4.664} & ~0.132          & ~\textbf{4.796} & ~9.03          & 25.00          & 29.17          & 33.33          & ~\textbf{4.261} & ~0.132          & ~\textbf{4.393} & 15.56          & 21.67          & 29.44          \\
  \hline
  \hline
  & \multicolumn{7}{c||}{S2: unseen handedness} & \multicolumn{7}{c}{S3: unseen grasping} \\
  \hline
  & success & \multicolumn{3}{c|}{mean accum time (s)} & \multicolumn{3}{c||}{failure (\%)} & success & \multicolumn{3}{c|}{mean accum time (s)} & \multicolumn{3}{c}{failure (\%)} \\
  & (\%) & exec & plan & total & contact & drop & timeout & (\%) & exec & plan & total & contact & drop & timeout \\
  \hline
  OMG Planner~\cite{wang:rss2020}       & \textbf{62.78} & ~8.275          & ~1.481          & ~9.755          & 30.56          & ~\textbf{3.89} & ~\textbf{2.78} & \textbf{69.00} & ~8.478          & ~1.588          & 10.066          & 23.00          & ~\textbf{4.00} & ~\textbf{4.00} \\
  Yang et al.~\cite{yang:icra2021}      & 62.50          & ~4.808          & ~\textbf{0.034} & ~\textbf{4.843} & 16.11          & 10.56          & 10.83          & 62.00          & ~\textbf{4.805} & ~\textbf{0.031} & ~\textbf{4.837} & 18.00          & ~9.00          & 11.00          \\
  GA-DDPG~\cite{wang:corl2021} hold     & 55.00          & ~7.145          & ~0.129          & ~7.274          & ~\textbf{8.61} & 17.78          & 18.61          & 50.00          & ~7.305          & ~0.135          & ~7.440          & ~\textbf{5.00} & 23.00          & 22.00          \\
  GA-DDPG~\cite{wang:corl2021} w/o hold & 28.33          & ~\textbf{4.747} & ~0.133          & ~4.881          & ~9.17          & 34.44          & 28.06          & 33.00          & ~4.948          & ~0.123          & ~5.071          & 10.00          & 33.00          & 24.00          \\
  \hline
 \end{tabular}
 \caption{\small Performance of the baselines with our adopted metrics
(Sec.~\ref{sec:benchmark}) on the four evaluation setups.}
 \label{tab:quantitative}
\end{table*}

 \item \textbf{Yang et al.~\cite{yang:icra2021}.} This is a reactive H2R
handover system which performs point cloud-based grasp generation followed by
task planning for grasp selection. Since we assume ground-truth object pose,
we bypass grasp generation and directly use the pre-generated grasps as in the
OMG Planner for the task planning step.
 \item \textbf{GA-DDPG~\cite{wang:corl2021}.} This is a neural network policy
trained with RL for grasping static objects. It takes in a segmented point
cloud of the target object and directly outputs the robot's target joint 
position. The policy is closed-loop as the network is ran every 0.15 seconds.
To obtain the input, we render a point cloud from a wrist-mounted camera
following~\cite{wang:corl2021} and segment the point cloud using the
ground-truth segmentation mask. Once the gripper reaches the object, we switch
to the same hand coded policy as in the OMG Planner. We use the model trained
in~\cite{wang:corl2021}. Since it is trained only for grasping static objects, 
we evaluate this baseline with two variants: holding still until the human hand
comes to a stop as in the OMG Planner (``hold'') and without any hold (``w/o
hold'').
\end{itemize}
We note that none of these baselines has been trained on the benchmark. Our aim
is to first provide the results and analysis of existing approaches and models.
With proper baselines, our benchmark can pave the way for future handover
systems benefited from training in the environment.

\vspace{1mm}
\noindent \textbf{Results.} Tab.~\ref{tab:quantitative} shows the results on
the test splits of the four setups (i.e., S0, S1, S2, and S3). Below we focus
the discussion on S0 (default), since the results on the other three setups
also show similar trends.

\begin{figure*}[t]
 \centering
 \small
 \includegraphics[width=0.117\linewidth]{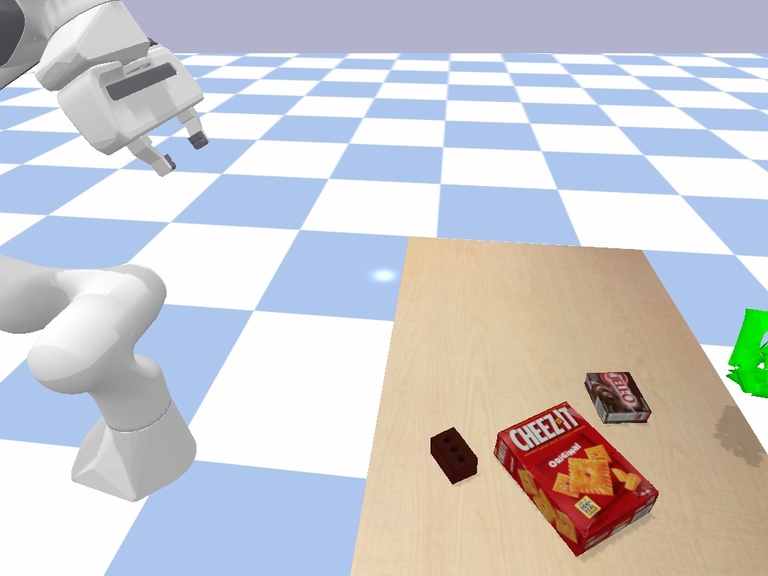}
 \includegraphics[width=0.117\linewidth]{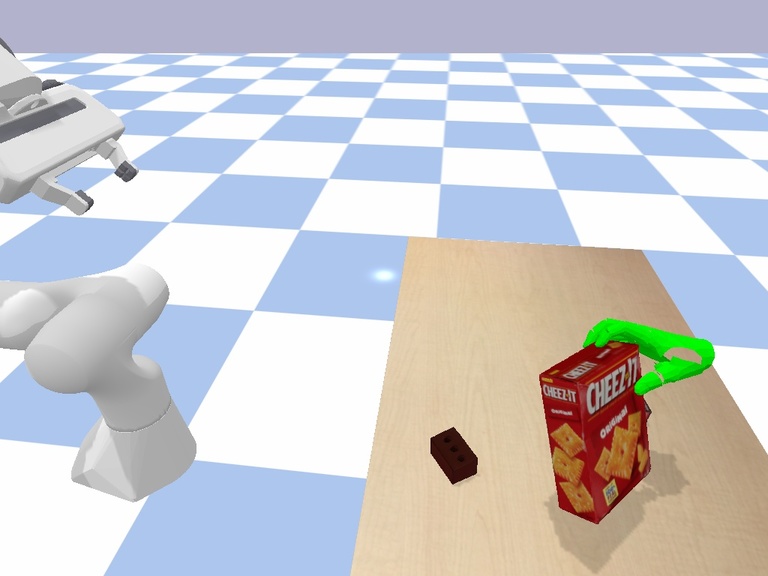}
 \includegraphics[width=0.117\linewidth]{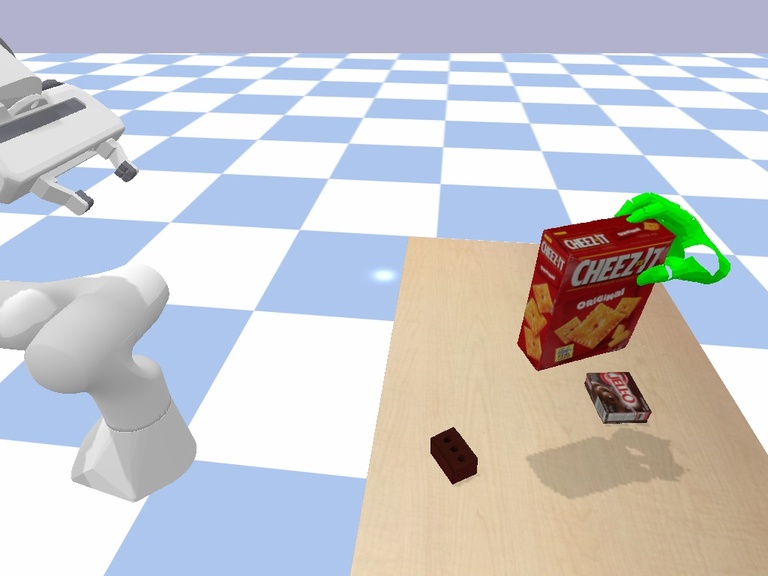}
 \includegraphics[width=0.117\linewidth]{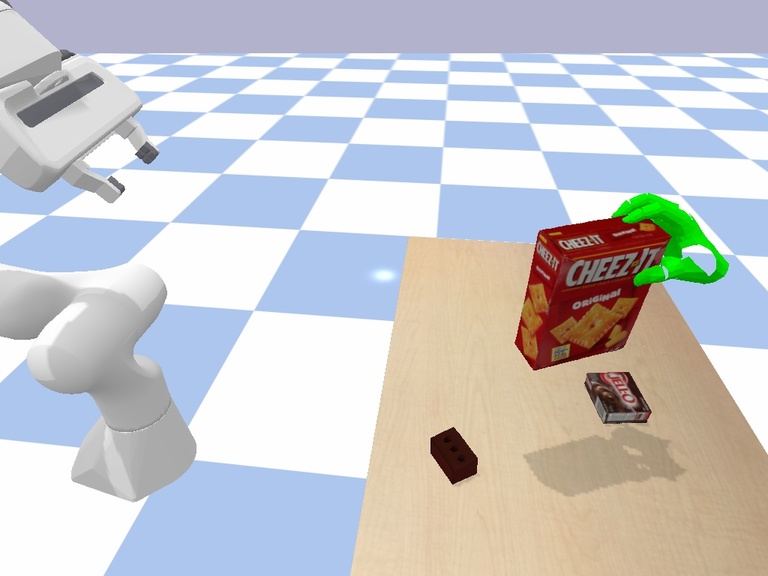}
 \includegraphics[width=0.117\linewidth]{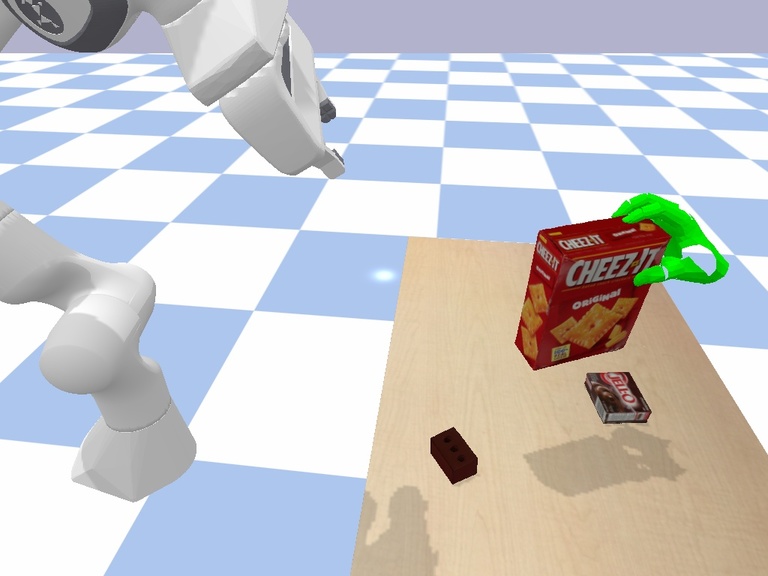}
 \includegraphics[width=0.117\linewidth]{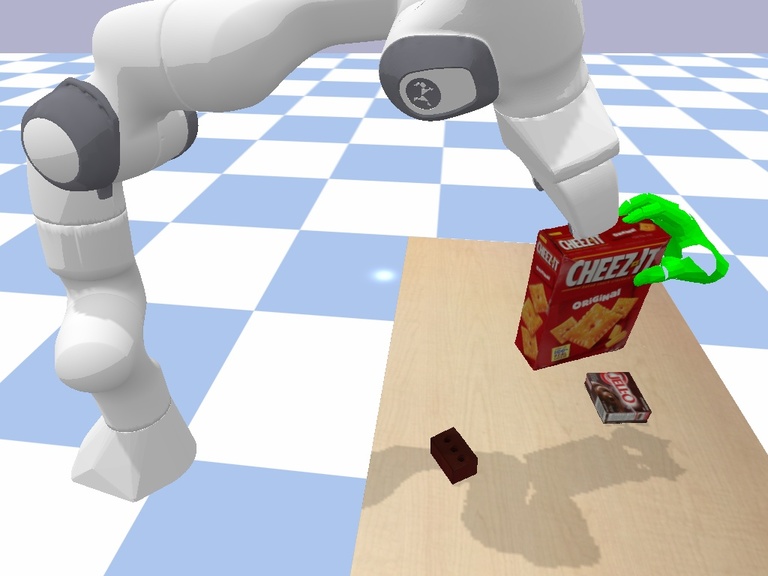}
 \includegraphics[width=0.117\linewidth]{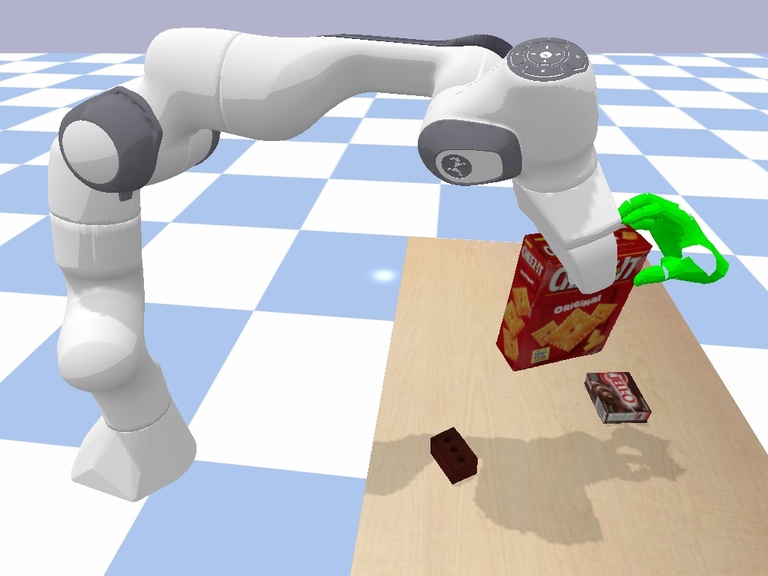}
 \includegraphics[width=0.117\linewidth]{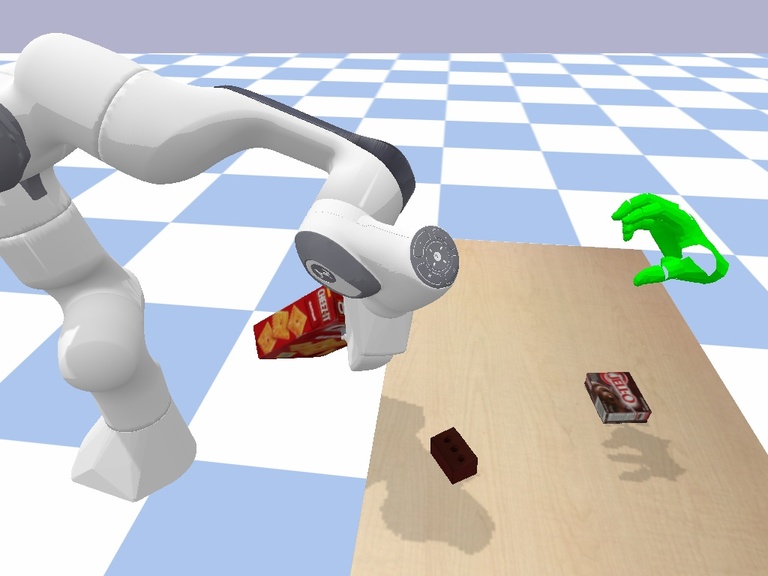}
 \\ \vspace{1mm}
 \includegraphics[width=0.117\linewidth]{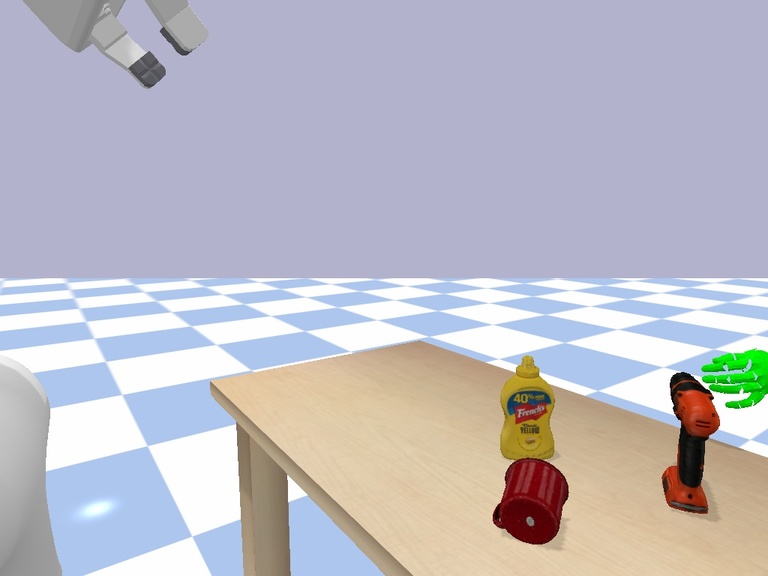}
 \includegraphics[width=0.117\linewidth]{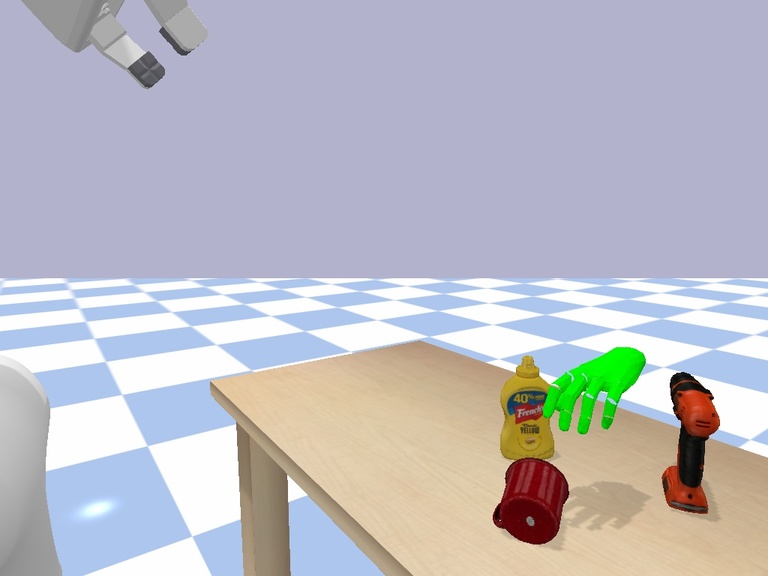}
 \includegraphics[width=0.117\linewidth]{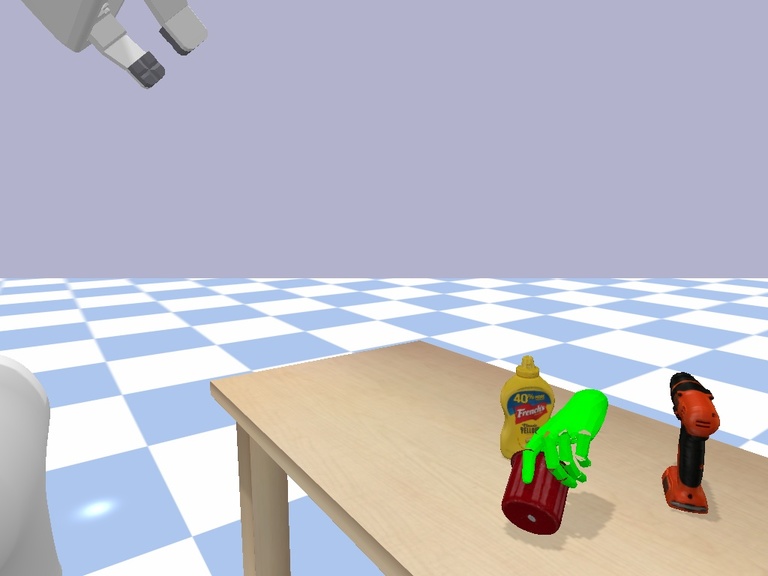}
 \includegraphics[width=0.117\linewidth]{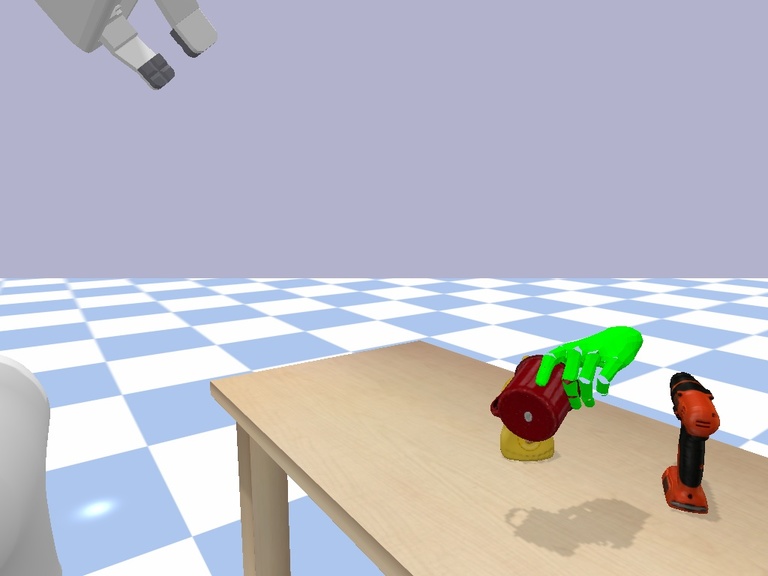}
 \includegraphics[width=0.117\linewidth]{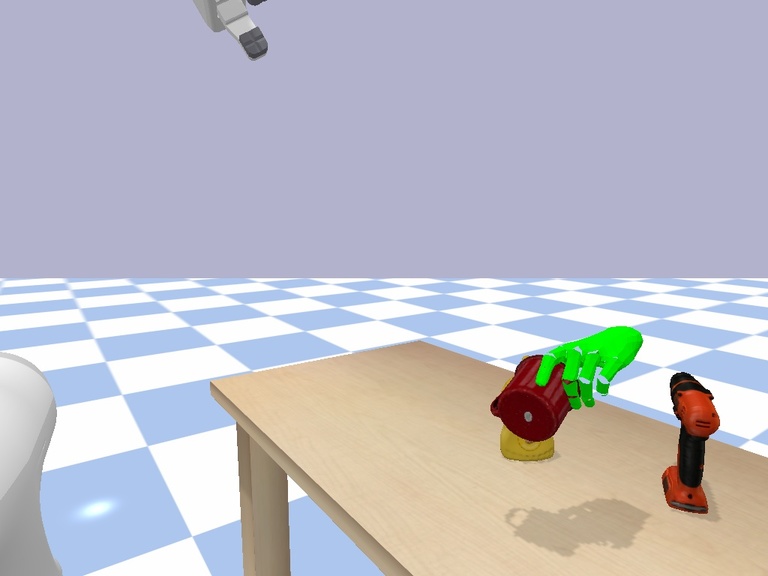}
 \includegraphics[width=0.117\linewidth]{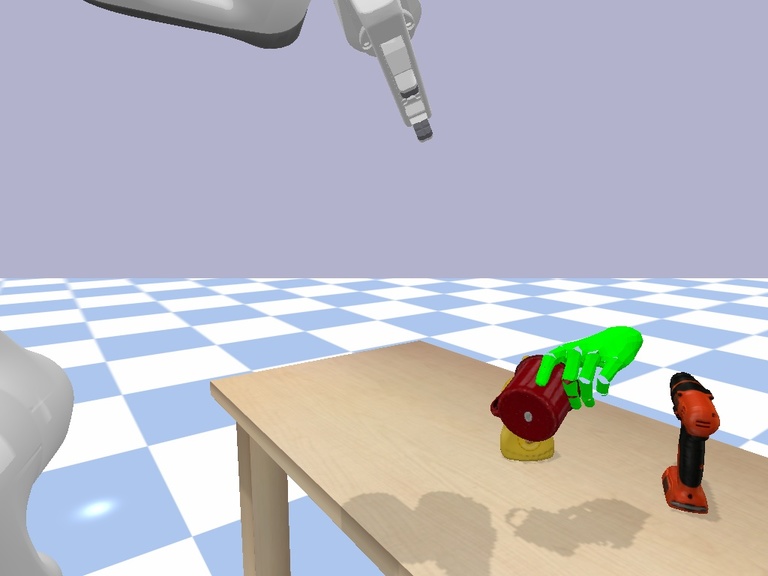}
 \includegraphics[width=0.117\linewidth]{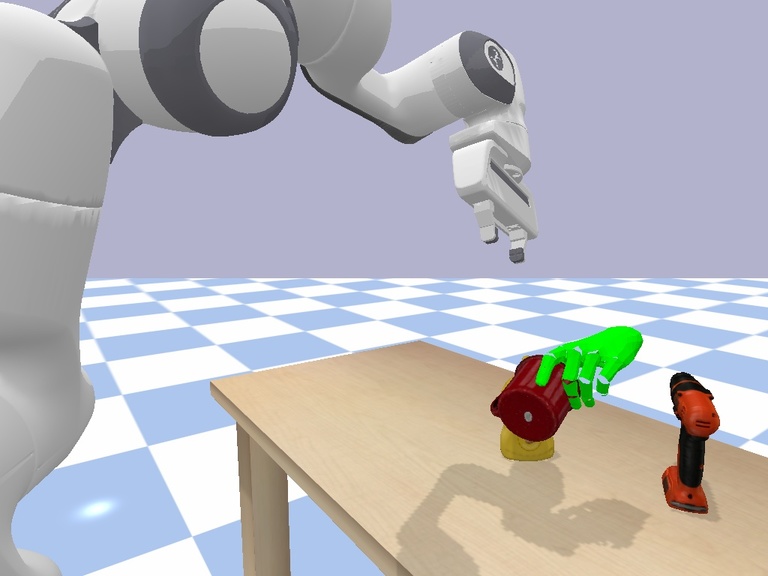}
 \includegraphics[width=0.117\linewidth]{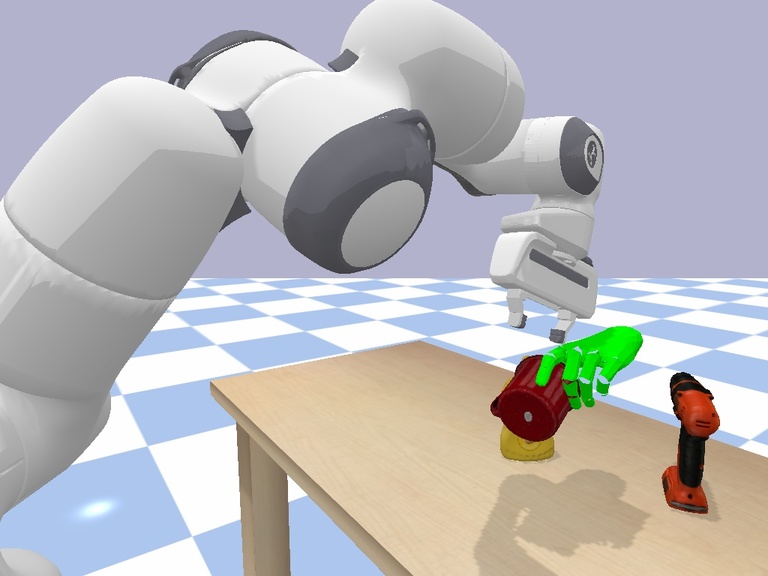}
 \\ OMG Planner~\cite{wang:rss2020} \\ \vspace{2mm}
 \includegraphics[width=0.117\linewidth]{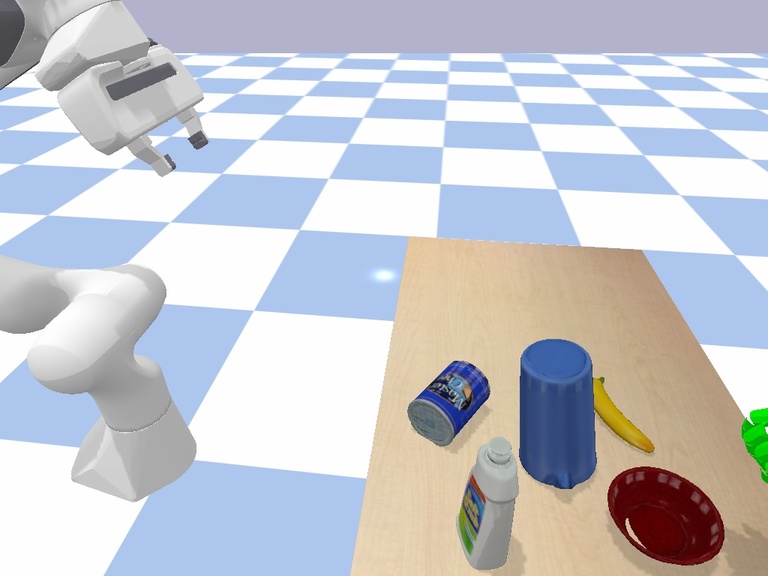}
 \includegraphics[width=0.117\linewidth]{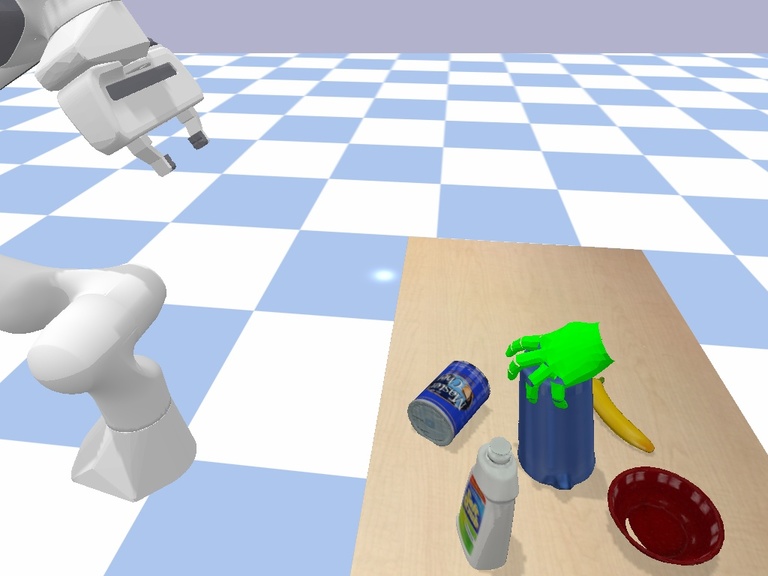}
 \includegraphics[width=0.117\linewidth]{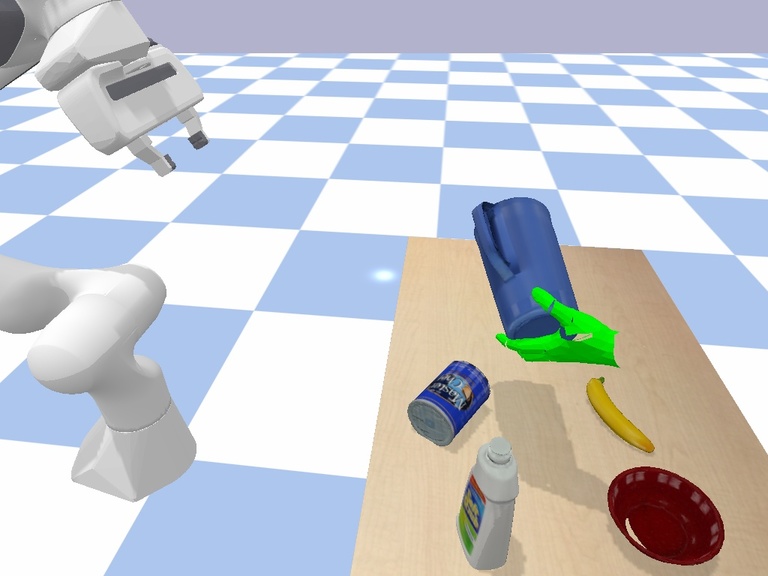}
 \includegraphics[width=0.117\linewidth]{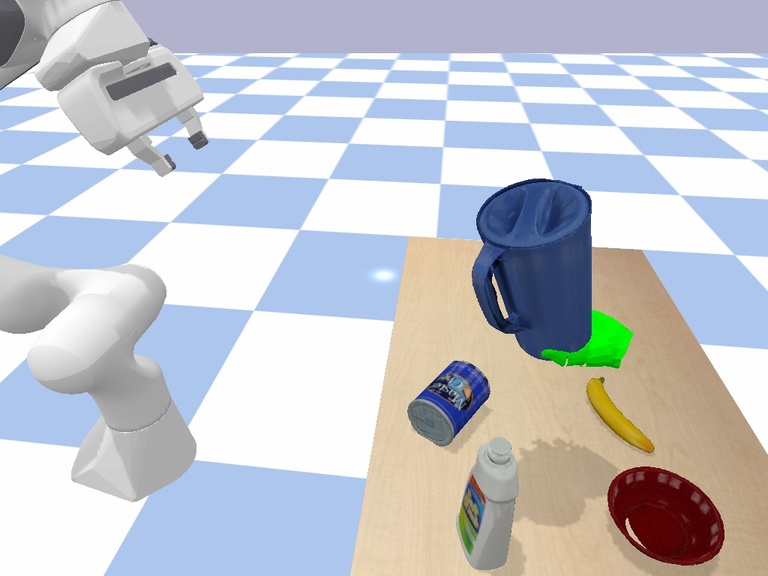}
 \includegraphics[width=0.117\linewidth]{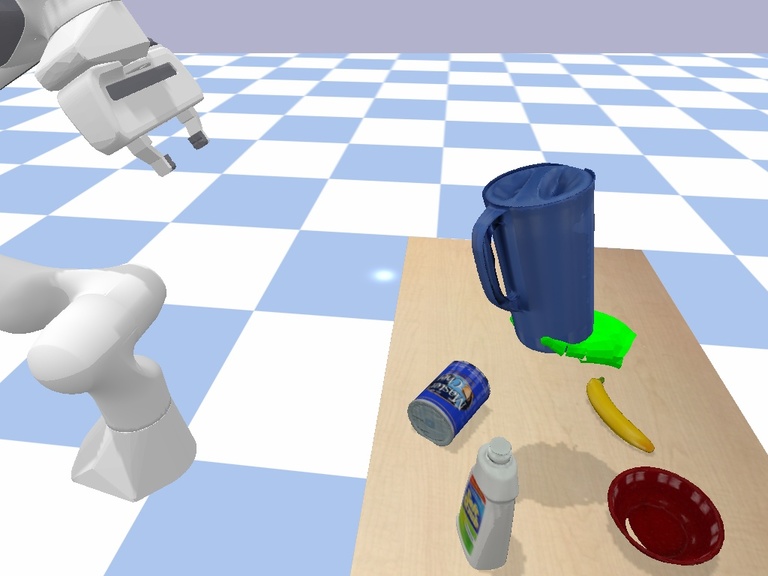}
 \includegraphics[width=0.117\linewidth]{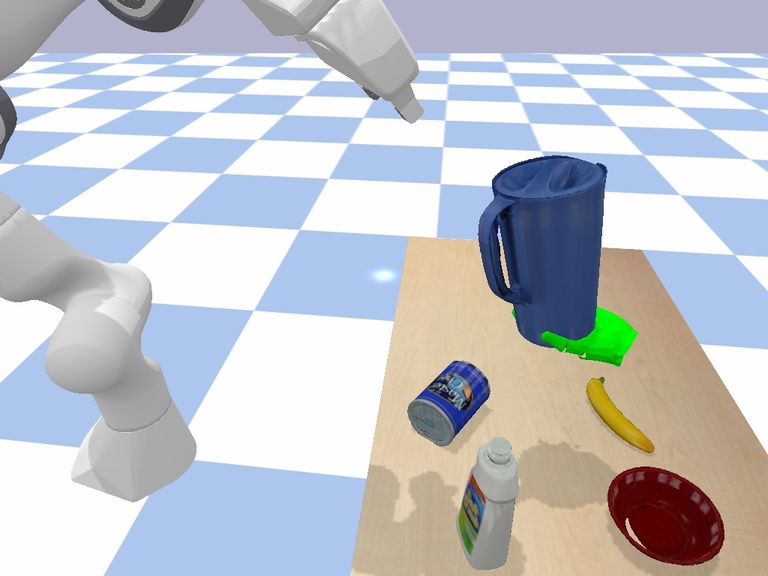}
 \includegraphics[width=0.117\linewidth]{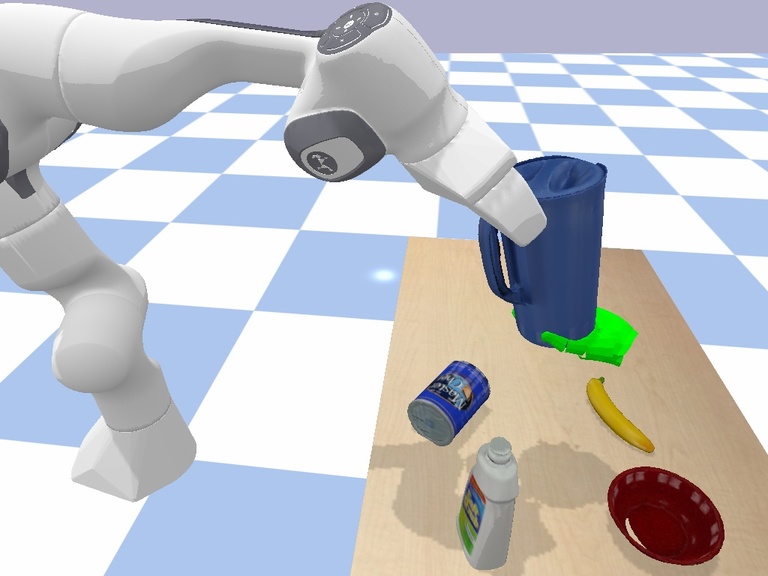}
 \includegraphics[width=0.117\linewidth]{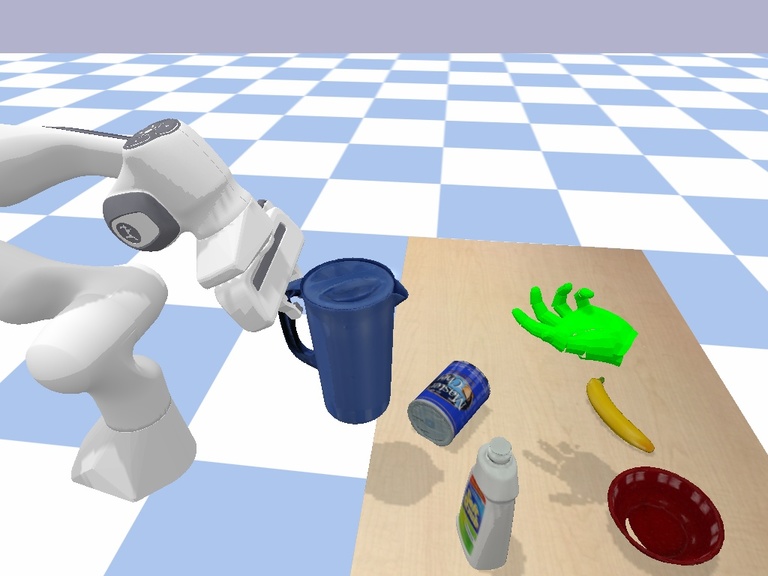}
 \\ \vspace{1mm}
 \includegraphics[width=0.117\linewidth]{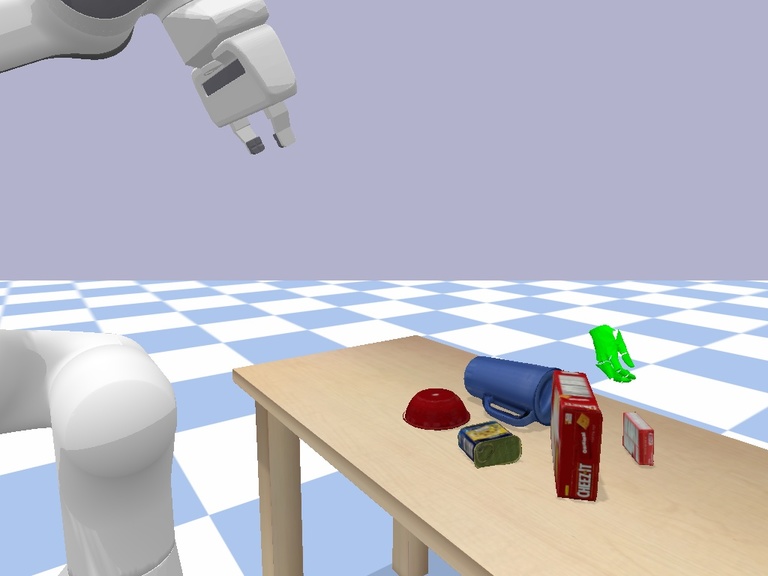}
 \includegraphics[width=0.117\linewidth]{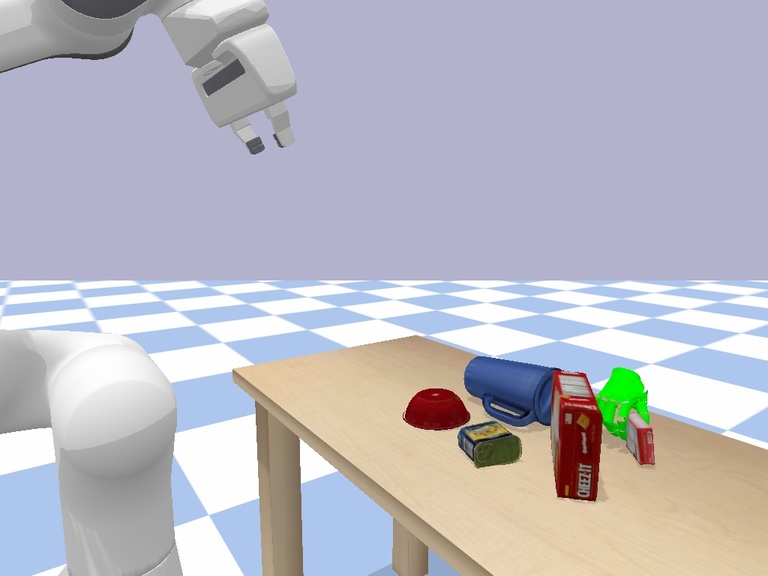}
 \includegraphics[width=0.117\linewidth]{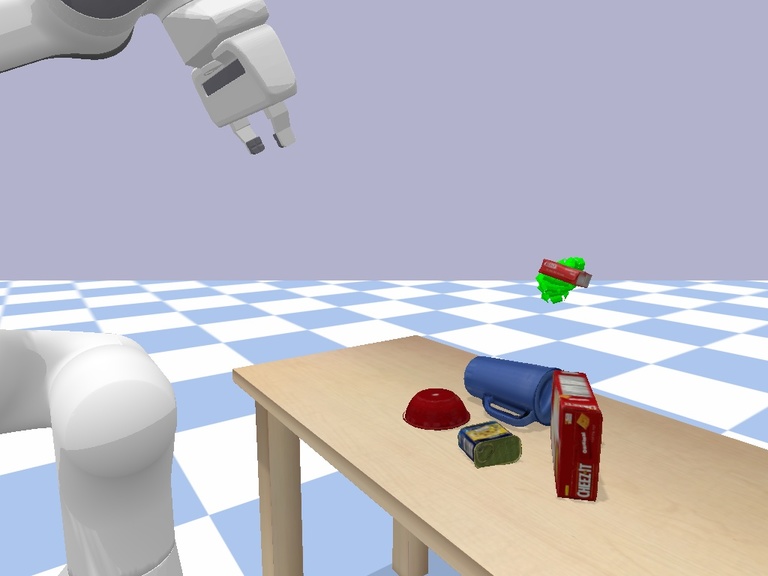}
 \includegraphics[width=0.117\linewidth]{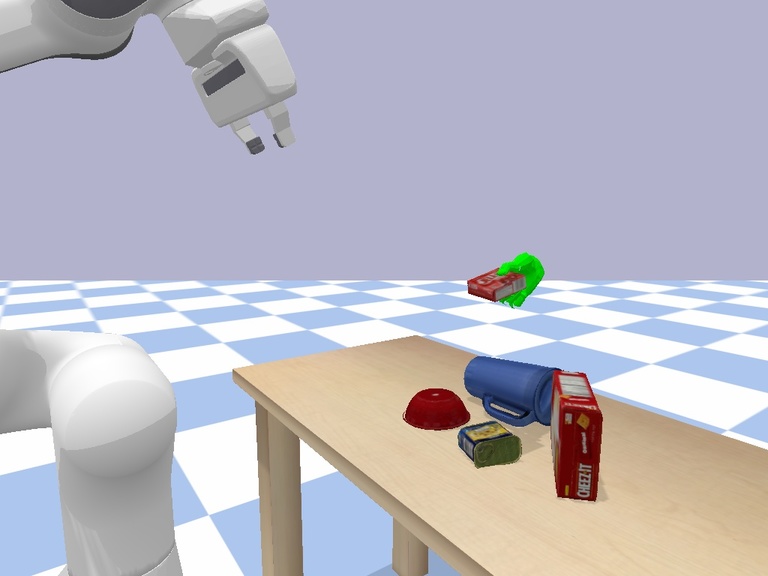}
 \includegraphics[width=0.117\linewidth]{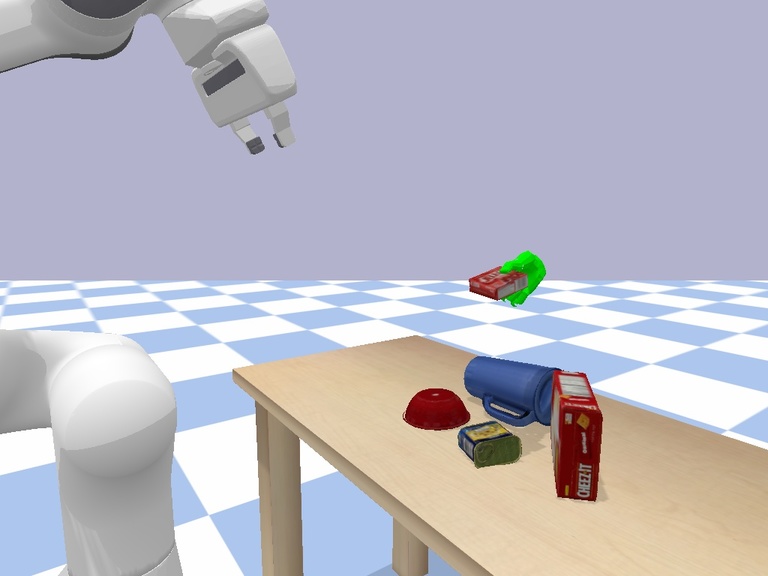}
 \includegraphics[width=0.117\linewidth]{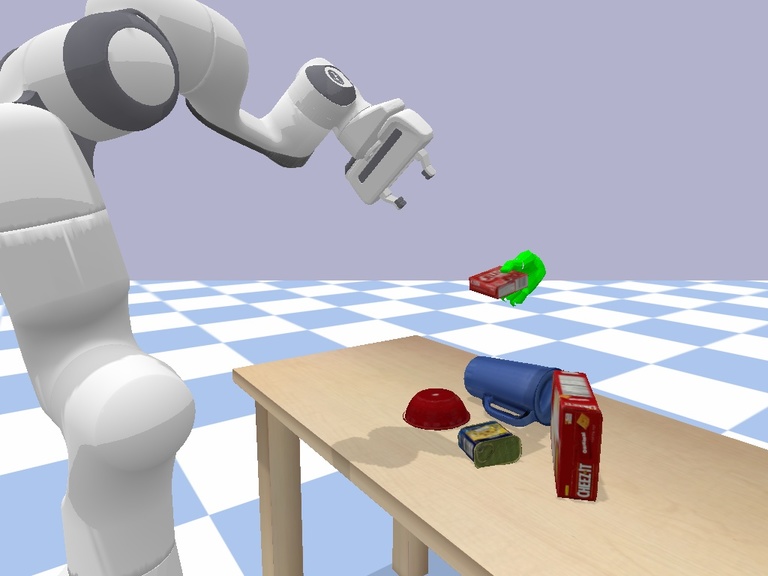}
 \includegraphics[width=0.117\linewidth]{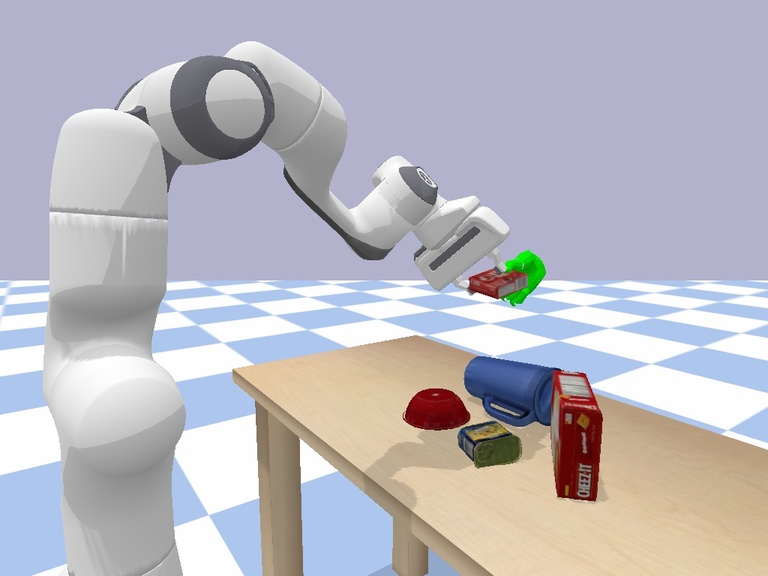}
 \includegraphics[width=0.117\linewidth]{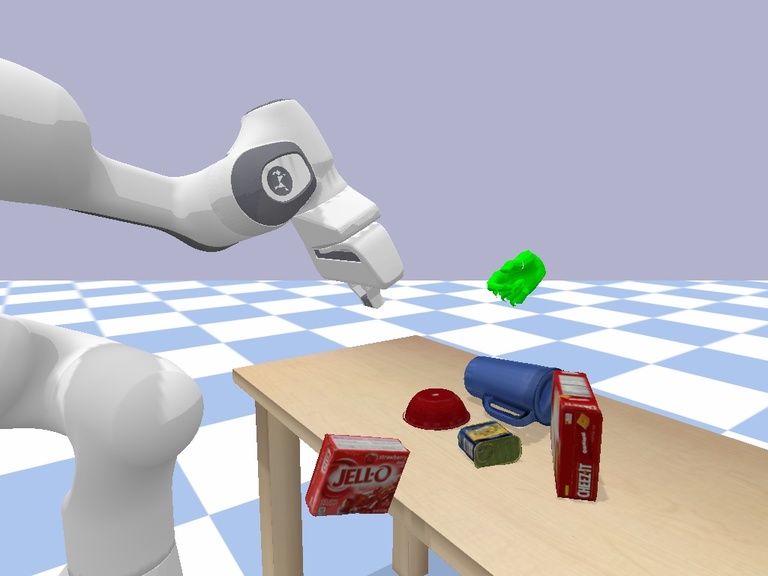}
 \\ Yang et al.~\cite{yang:icra2021} \\ \vspace{2mm}
 \includegraphics[width=0.117\linewidth]{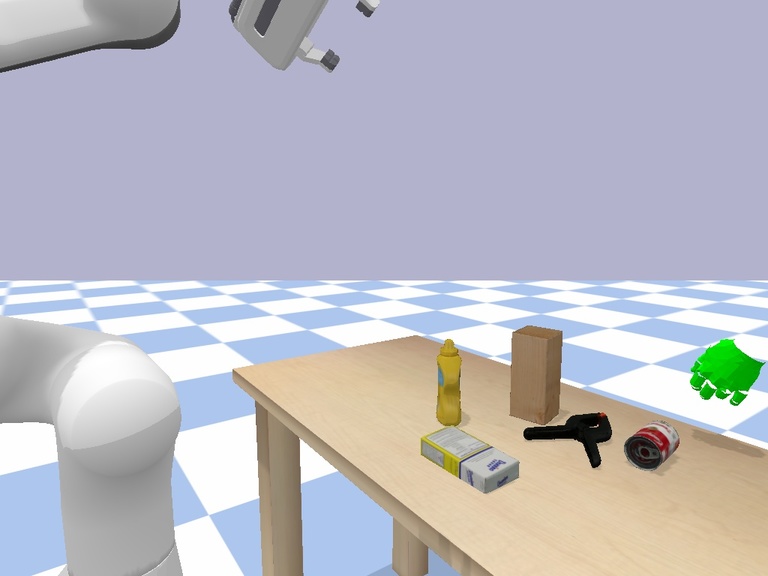}
 \includegraphics[width=0.117\linewidth]{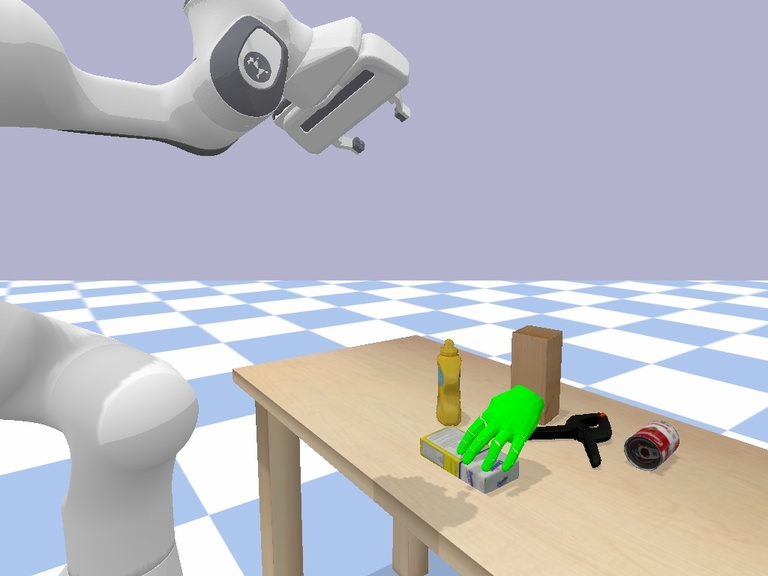}
 \includegraphics[width=0.117\linewidth]{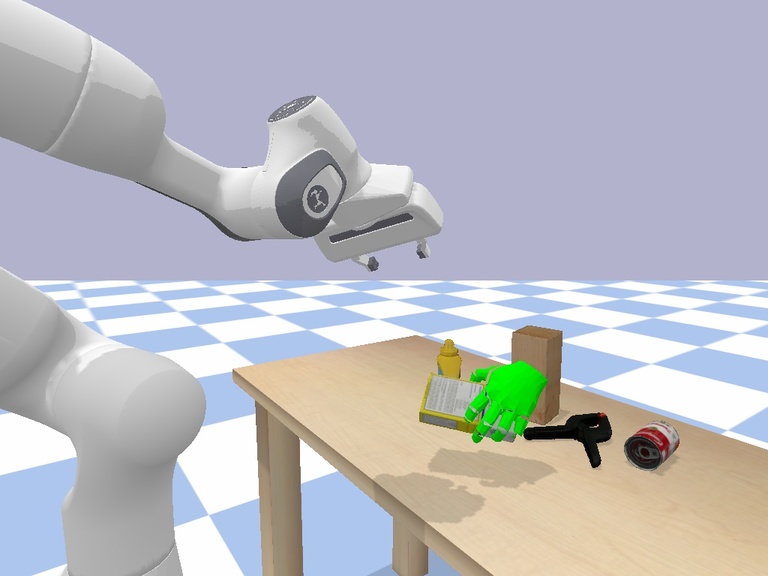}
 \includegraphics[width=0.117\linewidth]{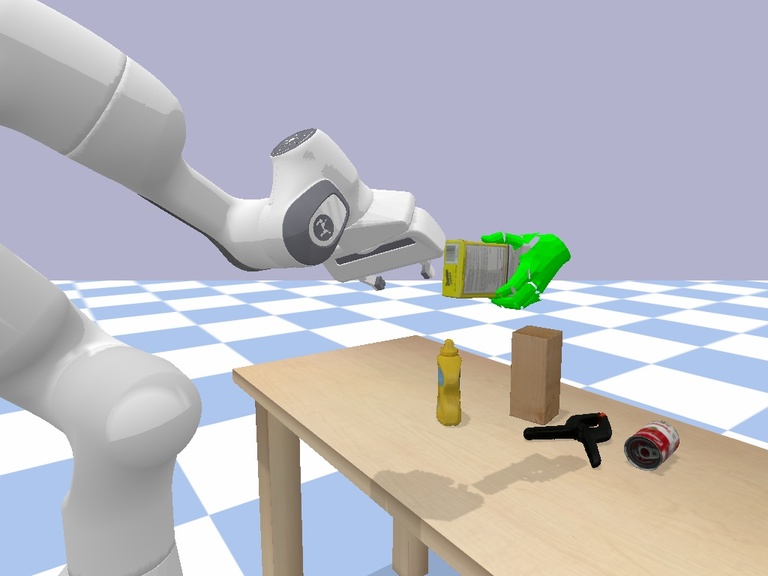}
 \includegraphics[width=0.117\linewidth]{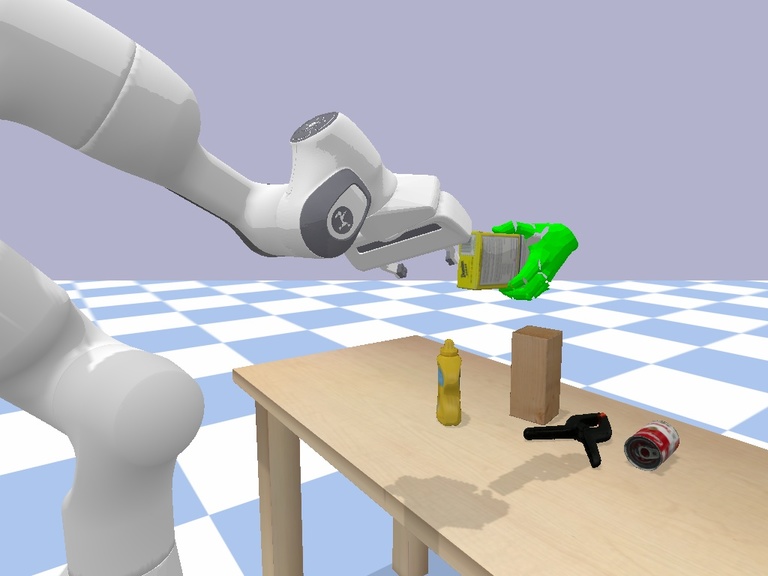}
 \includegraphics[width=0.117\linewidth]{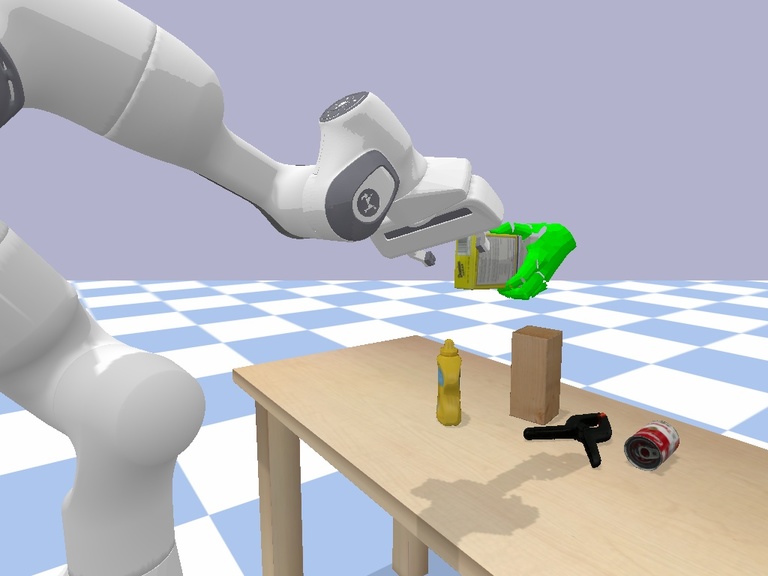}
 \includegraphics[width=0.117\linewidth]{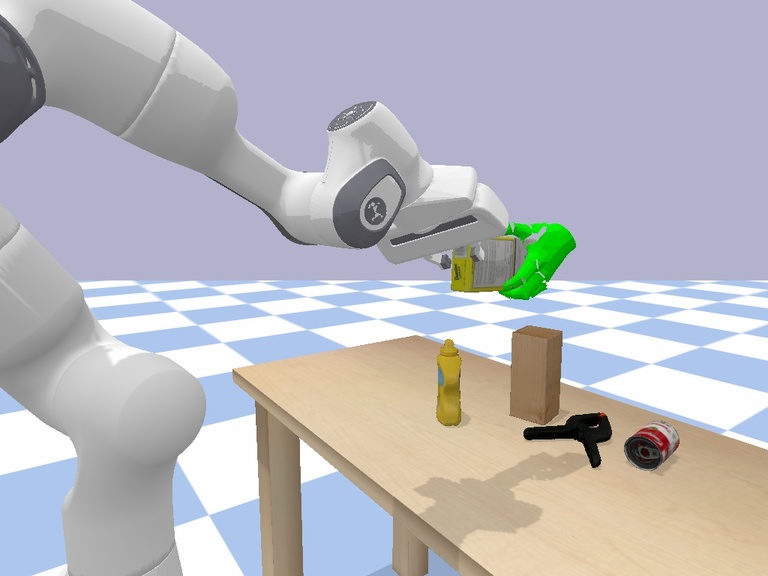}
 \includegraphics[width=0.117\linewidth]{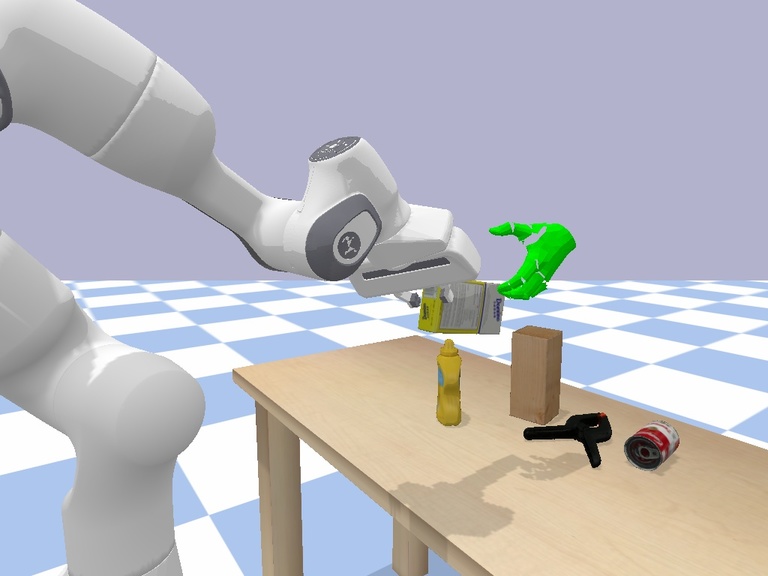}
 \\ \vspace{1mm}
 \includegraphics[width=0.117\linewidth]{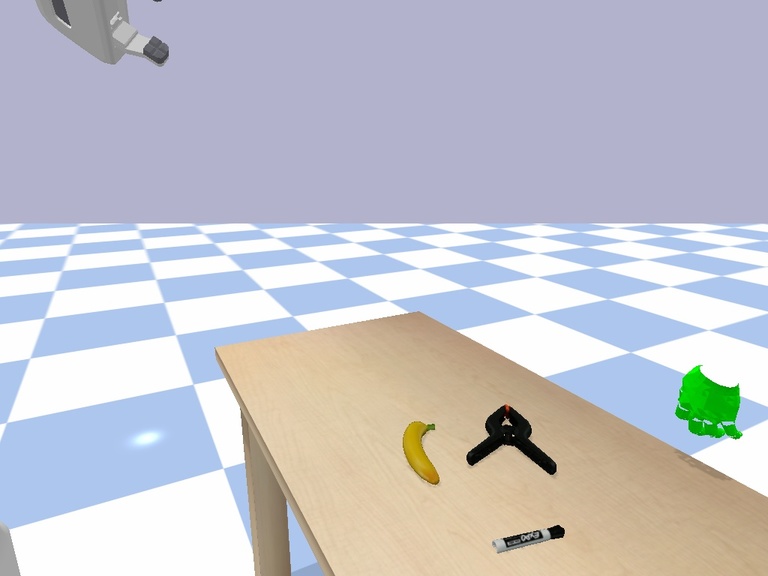}
 \includegraphics[width=0.117\linewidth]{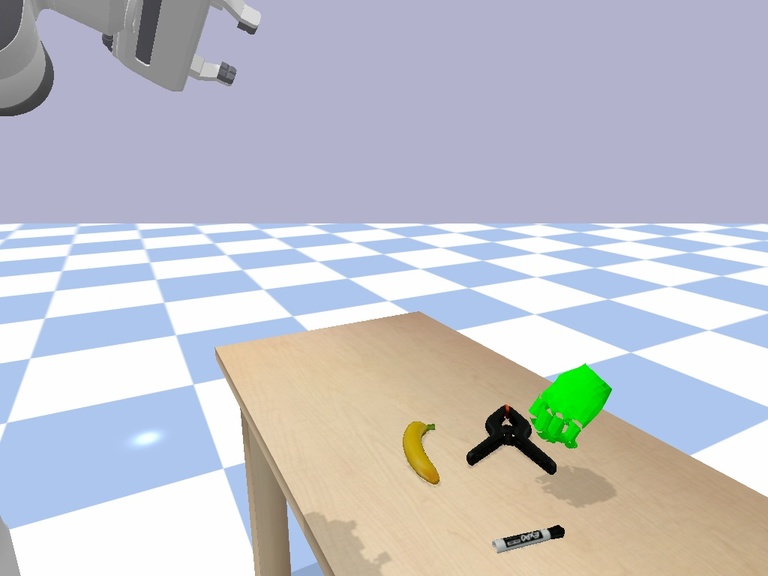}
 \includegraphics[width=0.117\linewidth]{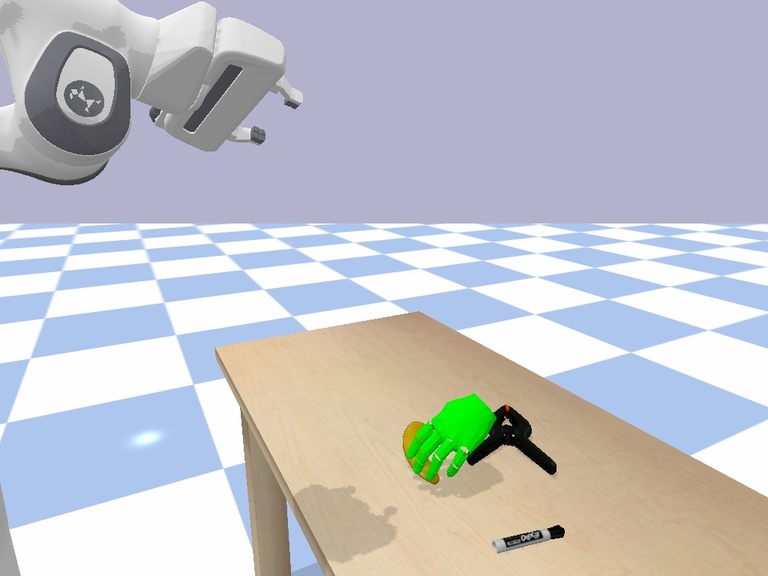}
 \includegraphics[width=0.117\linewidth]{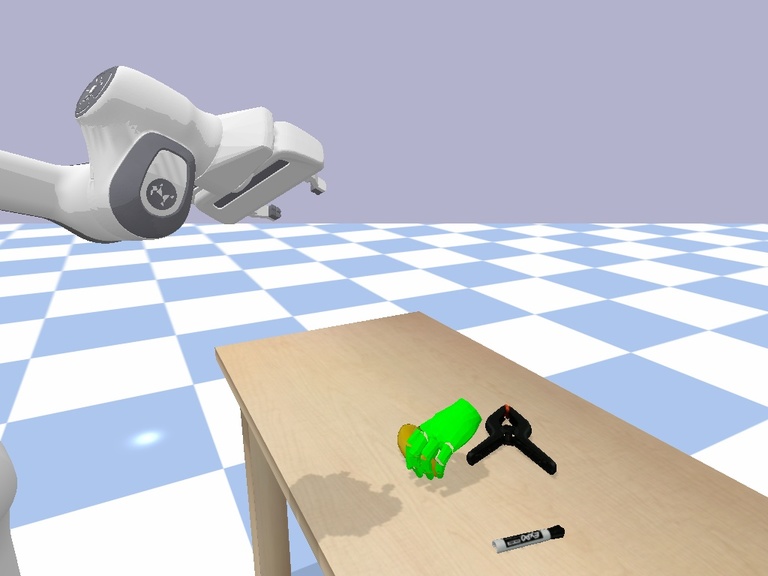}
 \includegraphics[width=0.117\linewidth]{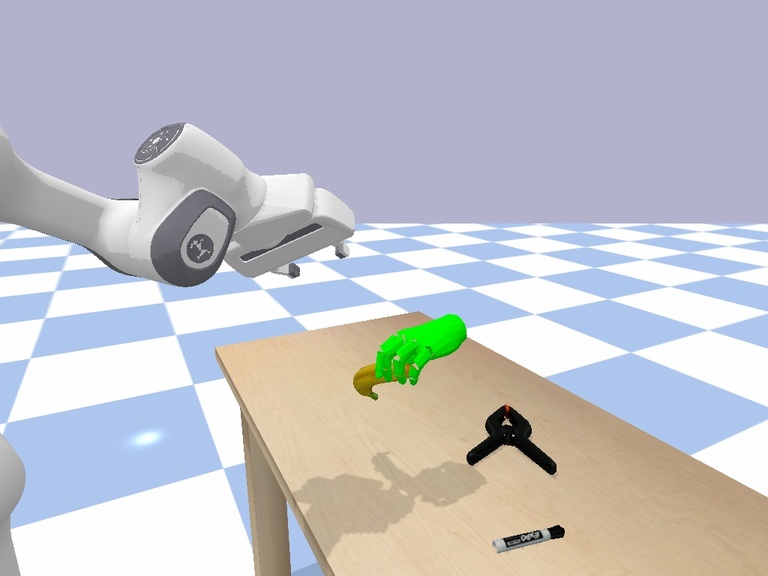}
 \includegraphics[width=0.117\linewidth]{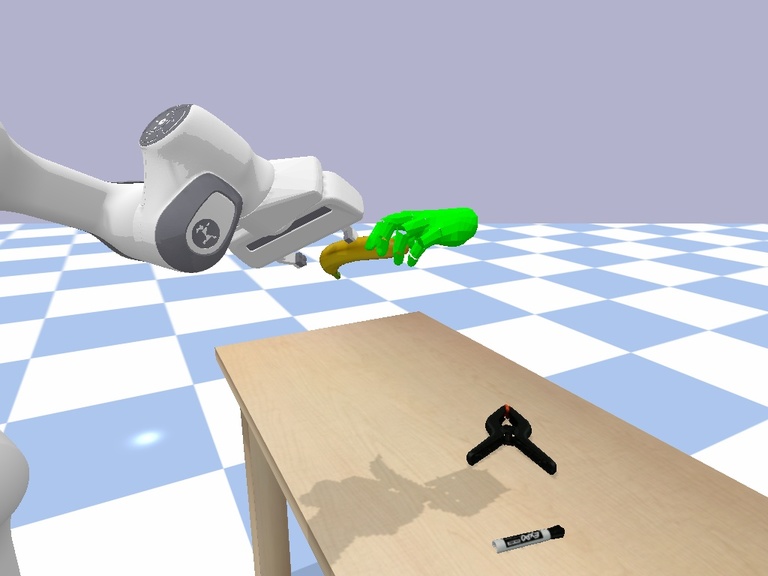}
 \includegraphics[width=0.117\linewidth]{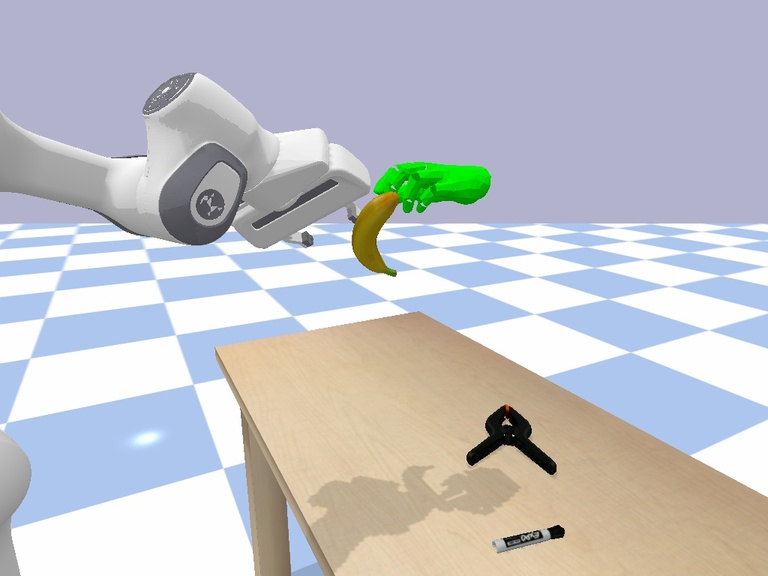}
 \includegraphics[width=0.117\linewidth]{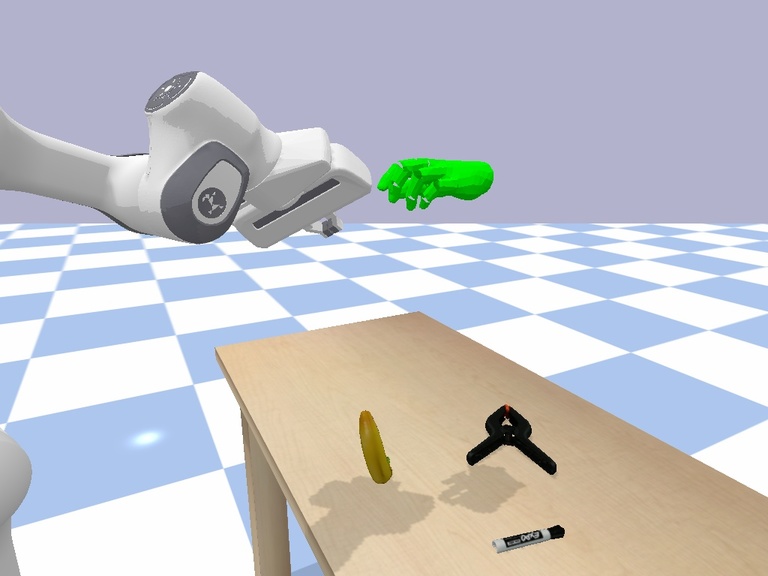}
 \\ GA-DDPG~\cite{wang:corl2021} w/o hold
 \caption{\small Qualitative results of handovers. For each baseline, we show
one success (top) and one failure (bottom) case.}
 \label{fig:qualitative}
\end{figure*}

We see that the OMG Planner achieves competitive success rate (62.50\%) among
all the baselines. Besides, the failure cases are dominated by robot-human
contact (27.28\%). This is unsurprising since the planner takes no account of
the hand's position and thus might generate hand-colliding trajectories.
However, object dropping (8.33\%) and timeout (1.39\%) both achieve the lowest
occurrence rate among all the baselines. This suggests that with accurate
object pose estimates and robust grasp generation, motion planning based
approaches can be very reliable. Despite the high efficacy, it falls short on
efficiency---the OMG Planner achieves the highest mean accumulated time among
all the baselines on both execution (8.309s) and planning (1.414s). The first
two rows in Fig.~\ref{fig:qualitative} shows qualitative examples of a success
(top) and a failure due to robot-human contact (bottom).

Yang et al. achieves a comparable success rate to the OMG Planner (64.58\%
versus 62.50\%). Yet the occurrence rate of robot-human contact is
significantly lower (17.36\% versus 27.78\%). This is attributed to the
system's tendency of approaching the object directly from the front side, which
is typically free from collision with the hand that holds the object from the
opposite side. Since the system is tailored for reactive handovers, the robot
is moving with higher acceleration and deceleration. This causes the object to
drop more easily during contact, resulting in a higher failure rate of ``drop''
(11.81\% versus 8.33\%). However, the higher peak speed also improves the
efficiency. The system achieves competitive mean accumulated time on execution
(4.864s), only behind the ``w/o hold'' variant of GA-DDPG. The middle two rows
in Fig.~\ref{fig:qualitative} shows qualitative examples of a success (top) and
a failure due to object dropping (bottom).

GA-DDPG (hold) achieves a lower success rate compared to the OMG Planner and 
Yang et al. (50.00\% versus 62.50\% and 64.58\%). This is due to the more
ambiguous point cloud input compared to ground-truth object pose and
pre-generated grasps. We also see a much higher rate on timeout (e.g., 25.69\%
versus 1.39\%), many of which are resulted from grasping in wrong locations. In
contrast, the failure rate due to contacting human is much lower (e.g., 4.86\%
versus 27.78\%), since mis-grasps often happen even before the gripper gets
close enough to the hand. In terms of efficiency, the execution time is
comparable to Yang et al. (4.664 versus 4.864 seconds), but planning is
slightly slower (0.142 versus 0.036 seconds) due to additional point cloud
processing.

Finally, GA-DDPG (w/o hold) achieves the lowest success rate among all the
baselines (36.81\%). The failures are often caused by the gripper contacting
the human hand or object when the hand and object are still actively moving,
since the policy was never trained to adapt to moving objects. However, this
baseline achieves the lowest mean accumulated time on execution and total,
since it is not forced to hold. The last two rows in Fig.~\ref{fig:qualitative}
shows qualitative examples of a success (top) and a failure from knocking down
the object (bottom) from GA-DDPG (w/o hold).

\vspace{1mm}
\noindent \textbf{Correlation with Real-World Evaluation.} A critical question
for a simulation benchmark is whether the achieved performance translates into
real-world performance. Below we refer to a relevant real-world user study
reported in Sec. 4.3 of~\cite{wang:corl2021}. The study compares Yang et
al.~\cite{yang:icra2021} and GA-DDPG~\cite{wang:corl2021} on H2R handover with
6 participants. We excerpt the results in Tab.~\ref{tab:real}.

We observe a positive correlation between the performance in the real world and
on HandoverSim. In terms of the success rate, Yang et al. has a slight edge
over GA-DDPG in Tab.~\ref{tab:real} (i.e., 82\% versus 80\%), and the same
trend also holds on HandoverSim (e.g., 62.78\% versus 55.00\% on S1). In terms
of efficiency, Yang et al. achieves a lower approach time than GA-DDPG in
Tab.~\ref{tab:real} (i.e., 10.7 versus 12.7 seconds), which again holds for the
mean accumulated time on HandoverSim (e.g., 4.758 versus 6.927 seconds).

In a subjective evaluation (Fig. 5 (bottom) in~\cite{wang:corl2021}), the users
are asked to score each given statement from 1 (strongly disagree) to 5
(strongly agree). The scores for the statement \textit{``The robot and I worked
fluently as a team to transfer objects.''} for GA-DDPG are (4, 4, 4, 4, 4, 3),
while the scores for Yang et al. are (5, 5, 4, 4, 4, 3) (Fig. 10
in~\cite{yang:icra2021}). The higher average score of Yang et al. (i.e., 4.17
versus 3.83) also positively correlates with its better efficiency performance
on HandoverSim.

Despite the correlation, we also see an offset between the performance in real
and simulation. On one hand, the success rates in real are constantly higher
than on HandoverSim (e.g., for GA-DDPG, 80\% versus 55.00\% on S1). This is
because real users are often cooperative and can help adjust the pose of the
object to prevent grasping failures, making the system more error tolerant. On
the other hand, the efficiency performance in real is constantly lower than on
HandoverSim (e.g., for Yang et al., a 10.7 seconds approach time versus a 4.758
seconds accumulated time on S1). This offset can be attributed to two factors.
First, the reported time on HandoverSim does not contain any latency from
perception, since the baselines directly consume the ground-truth state
information. In contrast, the real world systems involve perception stacks
(e.g., human body tracking and hand segmentation in~\cite{yang:icra2021}).
Second,~\cite{yang:icra2021} uses an extra low level control module (Riemannian
Motion Policies) to achieve smooth motion. This adds additional latency to the
loop. In constrast, HandoverSim uses a simple PD controller without any
sophisticated control modules.

\begin{table}[t]
 \centering
 \small
 \setlength{\tabcolsep}{4pt}
 \begin{tabular}{l|cc}
  \hline
  & avg. success rate (\%) & avg. approach time (s) \\
  \hline
  Yang et al.~\cite{yang:icra2021} & \textbf{82} & \textbf{10.7} \\
  GA-DDPG~\cite{wang:corl2021}     & 80          & 12.1          \\
  \hline
 \end{tabular}
 \caption{\small Results of the real-world user study from~\cite{wang:corl2021}.}
 \label{tab:real}
\end{table}

%
%
%

\section{Conclusions}

We have introduced a new simulation benchmark for H2R handovers. We have
analyzed the performance of a set of baselines on our benchmark, and validated
its credibility by showing a correlation with a real-world user study.






{
\bibliographystyle{IEEEtran}
\bibliography{root}
}

\end{document}